\documentclass[5p]{elsarticle}

\usepackage{graphics} 
\usepackage{epsfig} 
\usepackage{amsmath} 
\usepackage{amssymb}  
\usepackage{bm}  
\usepackage{subcaption}
\usepackage[font=small,labelfont=small]{caption}
\usepackage[hyphens]{url}
\usepackage{hyperref}
\usepackage{wasysym}
\usepackage{color}
\usepackage[framemethod=TikZ]{mdframed}
\definecolor{graybox}{rgb}{.9,.9,.9}
\mdfsetup{roundcorner=4,linewidth=0,backgroundcolor=graybox,leftmargin=0em,rightmargin=0em,innerleftmargin=0.3em,innerrightmargin=0.3em,innertopmargin=0.3em,innerbottommargin=0.3em}

\newcommand{\etal}{\MakeLowercase{\textit{et al.\ }}}
\newcommand\trsp{{\scriptscriptstyle\top}}

\graphicspath{{Figures/}}

\journal{Robotics and Autonomous Systems}

\begin{document}

\begin{frontmatter}

\title{Tensor-variate Mixture of Experts \\for Proportional Myographic Control of a Robotic Hand}

\author[idiap]{No\'emie Jaquier\corref{mycorrespondingauthor} \fnref{currentaffiliation}}
\cortext[mycorrespondingauthor]{Corresponding author}
\fntext[currentaffiliation]{No\'emie Jaquier is now affiliated with the Karlsruhe Institute of Technology, Adenauerring 2, 76131 Karlsruhe, Germany}
\ead{noemie.jaquier@kit.edu}

\author[citec]{Robert Haschke}
\ead{rhaschke@techfak.uni-bielefeld.de}

\author[idiap]{Sylvain Calinon}
\ead{sylvain.calinon@idiap.ch}

\address[idiap]{Idiap Research Institute, Rue Marconi 19, CH-1920 Martigny, Switzerland}
\address[citec]{Center of Excellence - Cognitive Interaction Technology (CITEC), Bielefeld University, D-33619 Bielefeld, Germany}

\begin{abstract}
When data are organized in matrices or arrays of higher dimensions (tensors), classical regression methods first transform these data into vectors, therefore ignoring the underlying structure of the data and increasing the dimensionality of the problem. This flattening operation typically leads to overfitting when only few training data is available. 
In this paper, we present a mixture-of-experts model that exploits tensorial representations for regression of tensor-valued data. The proposed formulation takes into account the underlying structure of the data and remains efficient when few training data are available. Evaluation on artificially generated data, as well as offline and real-time experiments recognizing hand movements from tactile myography prove the effectiveness of the proposed approach.
\end{abstract}


\end{frontmatter}


\section{Introduction}
\label{sec:Intro}
The data collected by a robot are often naturally represented as matrices or tensors, i.e., generalization of matrices to arrays of higher dimensions \cite{Kolda09}. 
Examples include video streams \cite{Zhao14}, movements in multiple coordinate systems \cite{Calinon16JIST}, electroencephalography (EEG) \cite{Miwakeichi04,Washizawa10} or tactile myography (TMG) data \cite{Koiva15}. Most approaches described in the literature consist of reorganizing the elements of these tensors into vectors before applying learning algorithms based on linear algebra operating on vector spaces. This flattening operation ignores the underlying structure of the original data. Moreover, the dimensionality of the resulting problem dramatically increases, creating high computational and memory requirements. Finally, the number of model parameters to estimate in the learning method may become high, which constitutes an important issue in the cases where only few training data are available.

With the burst of multidimensional data available in various fields of research, important efforts were turned toward extending standard dimensionality reduction and learning techniques to tensor data. In this context, Zare \etal \cite{Zare18} proposed a review of tensor decomposition methods by dividing them into three categories of problems usually targeted by principal component analysis (PCA), namely low-rank tensor approximation, low-rank tensor recovery, and feature extraction. 
In particular, multilinear PCA (MPCA) \cite{Lu08} and weighted MPCA \cite{Washizawa10} were proposed to extract features from tensor objects as a preprocessing step for classification. Similarly, linear discriminant analysis was extended to multilinear discriminant analysis in \cite{Tao07} and factor analysis was adapted to tensor data in \cite{Tang13}. In the context of regression, Guo \etal \cite{Guo12} proposed generalizations of ridge regression (RR) and support-vector regression (SVR) methods to tensor data, where they showed the superiority of tensor-based algorithms over the vector-based algorithms in various applications. A similar extension of RR to tensor data was proposed in \cite{Zhou13} with an application in magnetic resonance imaging (MRI). Following a similar process, tensor-variate logistic regression (LR) was proposed in \cite{Hung13,Tan13} for the classification of multidimensional data. Moreover, kernel-based frameworks such as Gaussian processes (GPs) \cite{Zhao14} were also adapted to tensors \cite{Signoretto11}.

In this paper, we introduce a tensor-variate mixture-of-experts (TME) model for regression. Mixture-of-experts (ME) models, first introduced by Jacobs \etal \cite{Jacobs91}, combine the predictions of several experts based on their probability of being active in a given region of the input space. Each expert acts as a regression function, while a gate determines the regions of the input space where each expert is trustworthy. The output of the model is a weighted sum of the experts predictions. Over the years, ME models were widely improved with different gates, regression and classification models for the experts (see \cite{Yuksel12} for a review of applications). 
Notably, for wrist movements recognition based on electromyographic (EMG) signals, the ME model with linear experts can achieve similar performance as more complex nonlinear methods, at a lower computational cost \cite{Hahne14}. 

The contributions of this paper are two-fold: (\emph{i}) we propose a tensor-variate mixture-of-experts model that exploits tensorial representations to take into account the structure of the data in the regression process; and (\emph{ii}) we demonstrate the efficiency of tensor-based approaches in the context of prosthetic hands to recognize hand movements from tactile myography. In particular, we present a teleoperation experiment based on tactile myography to control a robotic arm and hand. 

In order to handle tensor data in a ME model, we propose to use tensor-variate models for both the experts and the gates. In Section~\ref{sec:TensorMixtExp}, we show that the experts can be defined as tensor linear models and that the gates can be set as tensor-variate softmax functions. Both elements are based on the inner product between the input tensor data and model parameters (presented in Section~\ref{sec:Background}). The resulting TME model is trained with an EM algorithm using the CANDECOMP/PARAFAC (CP) decomposition, also called canonical polyadic decomposition \cite{Carroll70,Harshman70}. 

The functionality of the proposed approach is first evaluated and compared to the corresponding vector-based approach using artificially generated data (Sections~\ref{subsec:Experiment2d}, ~\ref{subsec:ExperimentHd}). The TME model is then exploited in the context of prosthetic hands to recognize hand movements from tactile myography (TMG). 

Our TMG sensor is made of 320 cells organized in a $8\times40$ array forming a bracelet \cite{Koiva15}, therefore providing intrinsically matrix-valued data. 
Despite our data contains patterns that could be treated by deep learning strategies such as convolutional neural networks, the use of such approaches would require large training datasets to be efficient, which does not fit with the requirement of our application, targeting personalized calibration of prosthetic hands from very small datasets. 

The effectiveness of our approach is tested in an offline experiment with the aim of detecting finger and wrist movements from TMG data (Section~\ref{subsec:ExperimentTmg}). We show that the TME model outperforms the standard ME model and achieves similar performance as a GP at a lower computational cost, with the advantage of being easily interpretable due to the tensor structure. 
We finally validate the use of the proposed approach in a real-time teleoperation experiment, where participants controlled a robotic arm and hand by moving their wrist and closing/opening their hand (Section~\ref{subsec:ExperimentTeleoperation}).

\section{Preliminaries}
\label{sec:Background}
Tensors are generalization of matrices to arrays of higher dimensions \cite{Kolda09}, where vectors and matrices correspond to 1st and 2nd-order tensors. Tensor representation permits to represent and exploit the intrinsic structure of multidimensional arrays. We introduce here the tensor operations necessary for the proposed TME, as well as the extensions of two forms of the generalized linear model, namely ridge regression and logistic regression, to tensor-variate data.

\begin{table*}[!t]   
\begin{mdframed}
	\caption{Tensor notations and operations.}
	\label{Table:TensorOps}
	\renewcommand*{\arraystretch}{1.5}
		\begin{tabular}{c  p{8.0cm}}	
			Variable / operation & Description \\	
			\hline
			$M$ & Number of dimensions or modes of a tensor \\
			$\bm{x}\in\mathbb{R}^{I_1}$ & Vector variable \\
			$\bm{X}\in\mathbb{R}^{I_1\times I_2}$ & Matrix variable \\
			$\bm{\mathcal{X}}\in\mathbb{R}^{I_1\times \dots \times I_M}$ & Tensor variable \\
			$\bm{X}_{(m)} \in\mathbb{R}^{I_m\times (I_1 \ldots I_{m-1} I_{m+1} \ldots I_M )}$ & $m$-mode matricization or unfolding of a tensor\\
			\hline
			$(\bm{u}^{(1)} \circ \bm{u}^{(2)} \circ \ldots \circ \bm{u}^{(M)})_{i_1,i_2,\ldots,i_M} = u_{i_1}^{(1)}u_{i_2}^{(2)}\ldots u_{i_M}^{(M)}$ & Outer product\\
			$\bm{A}\odot\bm{B} = [\begin{matrix} \bm{a}_1\otimes\bm{b}_1 &\ldots & \bm{a}_K\otimes\bm{b}_K \end{matrix} ]\in\mathbb{R}^{IJ\times K}$ & Khatri--Rao product of two matrices $\bm{A}\in\mathbb{R}^{I\times K}$ and $\bm{B}\in\mathbb{R}^{J\times K}$\\
			$\bm{U}^{(m)} =[\bm{u}_1^{(m)} \ldots \bm{u}_R^{(m)}]$ & Factor matrix \\
			$\bm{U}^{(-m)} = ( \bm{U}^{(M)} \odot \ldots \odot \bm{U}^{(m+1)} \odot \bm{U}^{(m-1)} \odot \ldots \odot \bm{U}^{(1)})$ & Khatri--Rao products without factor matrix $\bm{U}^{(m)}$\\
			\hline
			
			$\bm{\mathcal{Y}} = \bm{u}^{(1)} \circ \bm{u}^{(2)} \circ \ldots \circ \bm{u}^{(M)}$ & Rank-one tensor \\
			$\bm{\mathcal{Y}} = \sum_{r=1}^{R} \bm{u}_{r}^{(1)} \circ \bm{u}_{r}^{(2)} \circ \ldots \circ \bm{u}_{r}^{(M)}$ & Rank-$R$ tensor \\
			$\bm{\mathcal{Y}} \approx \sum_{r=1}^{R} \bm{u}_{r}^{(1)} \circ \bm{u}_{r}^{(2)} \circ \ldots \circ \bm{u}_{r}^{(M)}$ & CP decomposition \\
			
			$\bm{Y}_{(m)} \approx \bm{U}^{(m)} \bm{U}^{(-m)\trsp}$ & $m$-mode matricization of the CP decomposition \\
			$\text{vec}(\bm{\mathcal{Y}}) \approx (\bm{U}^{(M)} \odot \ldots \odot \bm{U}^{(1)}) \bm{1}_{R}$ & Vectorization of the CP decomposition \\
			\hline
			
			$\langle \bm{\mathcal{X}}, \bm{\mathcal{Y}} \rangle = \sum_{i_1=1}^{I_1} \sum_{i_2=1}^{I_2} \ldots \sum_{i_M=1}^{I_M} x_{i_1,i_2,\ldots i_M} \; y_{i_1,i_2,\ldots i_M}$ & Inner product of two tensors $\bm{\mathcal{X}}$, $\bm{\mathcal{Y}}$ \\
			$\langle \bm{\mathcal{X}}, \bm{\mathcal{Y}} \rangle = \langle \bm{X}_{(m)}, \bm{Y}_{(m)} \rangle = \langle \text{vec}(\bm{\mathcal{X}}), \text{vec}(\bm{\mathcal{Y}}) \rangle$ & Inner product equivalences \\
			$\langle \bm{\mathcal{X}}, \bm{\mathcal{Y}} \rangle = \langle \bm{X}_{(m)},\; \bm{U}^{(m)} \bm{U}^{(-m)\trsp} \rangle \nonumber = \langle \bm{X}_{(m)} \bm{U}^{(-m)},\; \bm{U}^{(m)} \rangle$ & Inner product equivalence when $\bm{\mathcal{Y}}$ follows exactly a CP decomposition \\
		\end{tabular}
\end{mdframed}	
\end{table*}

\subsection{Tensor operations}
For the reader familiar with tensor methods, also called multilinear algebra, the tensor notations and operations introduced in the following are summarized in Table~\ref{Table:TensorOps}.

The inner product of two tensors $\bm{\mathcal{X}}$, $\bm{\mathcal{Y}}\in\mathbb{R}^{I_1\times I_2 \times \ldots \times I_M}$ is defined as the sum of the products of their entries, so that
\begin{equation}
\langle \bm{\mathcal{X}}, \bm{\mathcal{Y}} \rangle = \sum_{i_1=1}^{I_1} \sum_{i_2=1}^{I_2} \ldots \sum_{i_M=1}^{I_M} x_{i_1,i_2,\ldots i_M} \; y_{i_1,i_2,\ldots i_M}.
\label{Eq:TensorInnerProd}
\end{equation}

Note that the inner product of two tensors is equivalent to the Frobenius inner product of their $m$-mode matricization or unfolding $\bm{X}_{(m)}, \bm{Y}_{(m)} \in\mathbb{R}^{I_m\times (I_1 \ldots I_{m-1} I_{m+1} \ldots I_M )}$ and to the inner product of their vectorization $\text{vec}(\bm{\mathcal{X}})$,  $\text{vec}(\bm{\mathcal{Y}})\in\mathbb{R}^{I_1 \ldots I_M }$, namely 
\begin{equation}
\langle \bm{\mathcal{X}}, \bm{\mathcal{Y}} \rangle = \langle \bm{X}_{(m)}, \bm{Y}_{(m)} \rangle = \langle \text{vec}(\bm{\mathcal{X}}), \text{vec}(\bm{\mathcal{Y}}) \rangle .
\label{Eq:InnerProdEquivalence}
\end{equation}

A rank-one tensor $\bm{\mathcal{Y}}$ of order $M$ is a tensor that can be written as the outer product of $M$ vectors, i.e.,
\begin{equation}
\bm{\mathcal{Y}} = \bm{u}^{(1)} \circ \bm{u}^{(2)} \circ \ldots \circ \bm{u}^{(M)},
\label{Eq:RankOneTensor}
\end{equation}
with $\circ$ the outer product between vectors, so that each element of $\bm{\mathcal{Y}}$ is equal to $x_{i_1,i_2,\ldots,i_M} = u_{i_1}^{(1)}u_{i_2}^{(2)}\ldots u_{i_M}^{(M)}$ and $M$ the number of dimensions or modes of the tensor.

The CANDECOMP/PARAFAC (CP) decomposition (also called canonical polyadic decomposition) factorizes a tensor $\bm{\mathcal{Y}}\in\mathbb{R}^{I_1\times I_2 \times \ldots \times I_M}$ as a sum of $R$ rank-one tensors, i.e.,
\begin{equation}
\bm{\mathcal{Y}} \approx \sum_{r=1}^{R} \bm{u}_{r}^{(1)} \circ \bm{u}_{r}^{(2)} \circ \ldots \circ \bm{u}_{r}^{(M)}.
\label{Eq:CPdec}
\end{equation}
The smallest number of rank-one tensors that generate $\bm{\mathcal{Y}}$ as their sum is defined as the rank of the tensor $\bm{\mathcal{Y}}$. It corresponds to the smallest number of components $R=\text{rank}(\bm{\mathcal{Y}})$ in the CP decomposition.
The CP decomposition can also be expressed in terms of the $m$-mode matricization and vectorization of the tensor $\bm{Y}_{(m)}$ and $\text{vec}(\bm{\mathcal{Y}})$ as
\begin{align}
\bm{Y}_{(m)} &\approx \bm{U}^{(m)} \bm{U}^{(-m)\trsp} , 
\label{Eq:CPmat} \\
\text{vec}(\bm{\mathcal{Y}}) &\approx (\bm{U}^{(M)} \odot \ldots \odot \bm{U}^{(1)}) \bm{1}_{R},
\label{Eq:CPvec}
\end{align}	
where $\odot$ denotes the Khatri--Rao product, $\bm{U}^{(m)}\in\mathbb{R}^{I_m\times R}$ are factor matrices defined as
\begin{align*}
\bm{U}^{(m)} &=[\bm{u}_1^{(m)} \ldots \bm{u}_R^{(m)}], \\
\bm{U}^{(-m)} &= ( \bm{U}^{(M)} \odot \ldots \odot \bm{U}^{(m+1)} \odot \bm{U}^{(m-1)} \odot \ldots \odot \bm{U}^{(1)}),
\end{align*}
and $\bm{1}_{R}\in\mathbb{R}^{R}$ is a vector containing $R$ ones. The Khatri--Rao product of two matrices $\bm{A}\in\mathbb{R}^{I\times K}$ and $\bm{B}\in\mathbb{R}^{J\times K}$ results in a matrix $\bm{A}\odot\bm{B}\in\mathbb{R}^{IJ\times K}$ whose columns are equal to the Kronecker product of the columns of $\bm{A}$ and $\bm{B}$, i.e., $\bm{A}\odot\bm{B} = [\begin{matrix} \bm{a}_1\otimes\bm{b}_1 &\ldots & \bm{a}_K\otimes\bm{b}_K \end{matrix} ]$.

If $\bm{\mathcal{Y}}$ follows exactly a CP decomposition \eqref{Eq:CPdec}, the inner product \eqref{Eq:InnerProdEquivalence} can equivalently be written as
\begin{align}
\langle \bm{\mathcal{X}}, \bm{\mathcal{Y}} \rangle &= \langle \bm{X}_{(m)},\; \bm{U}^{(m)} \bm{U}^{(-m)\trsp} \rangle \nonumber \\ 
&= \langle \bm{X}_{(m)} \bm{U}^{(-m)},\; \bm{U}^{(m)} \rangle,
\label{Eq:InnerProdEquivalence2}
\end{align}
by exploiting \eqref{Eq:CPmat} and the properties of the Frobenius norm and matrix trace.

\subsection{Generalized linear model for tensors}
\begin{figure}[tbp]
	\centering
	\includegraphics[width=0.27\textwidth]{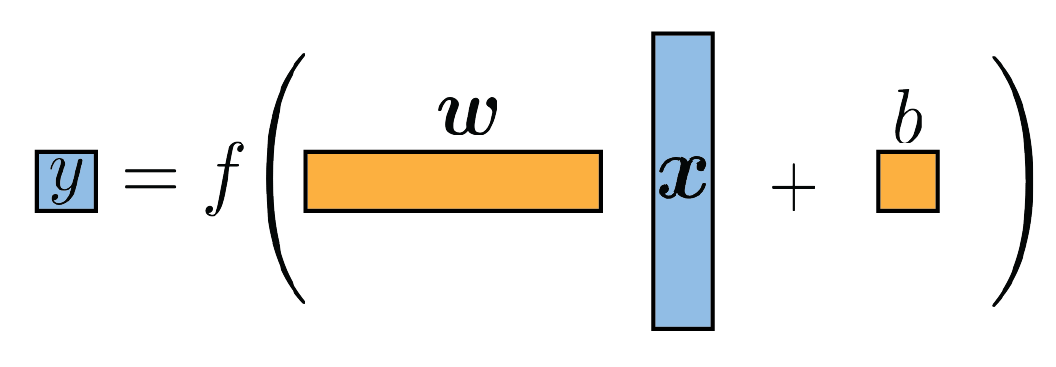}
	\includegraphics[width=0.32\textwidth]{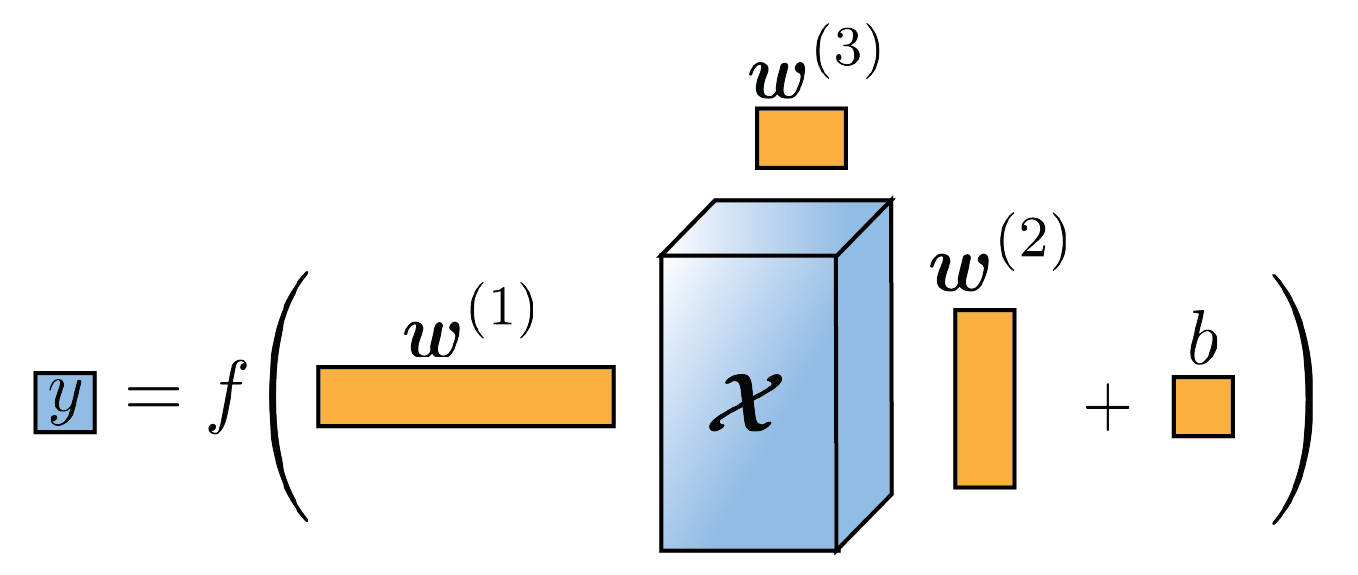}
	\includegraphics[width=0.5\textwidth]{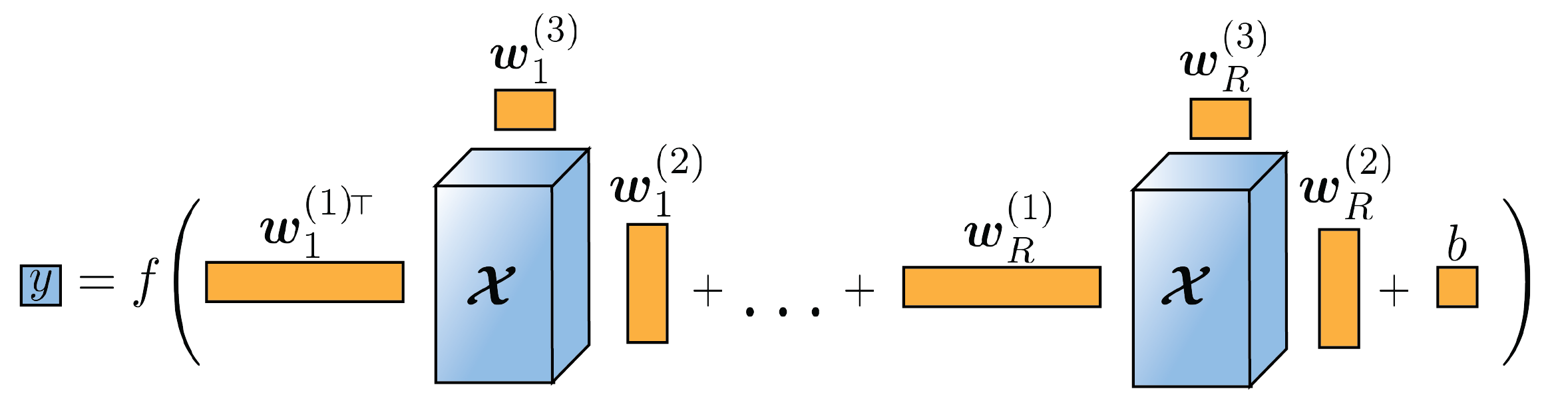}
	\caption{(\emph{Top}) GLM for vector-valued data. The data and model parameters are depicted in blue and orange, respectively. (\emph{Middle} and \emph{bottom}) Extensions of GLM to tensor-valued data. The \emph{bottom} representation allows the encoding of more complex behaviors as the weight tensor is encoded as a sum of rank-$1$ tensors.}  
	\label{Fig:GLRMmodels}
\end{figure}

Given a vector-valued input data $\bm{x}$, the generalized linear model (GLM) is given by
\begin{equation}
y = f(\bm{x}^\trsp\bm{w} + b) = f(\langle \bm{x}, \bm{w}\rangle +b),
\label{Eq:GLRM}
\end{equation}
where $y$ is the predicted output, $\bm{w}$ is a vector of weights, $b$ is the bias and $f(\cdot)$ is a function, see Figure~\ref{Fig:GLRMmodels}-\emph{top}.
This model can be naturally extended to matrix-valued data $\bm{X}$ with
\begin{align}
y &= f(\bm{w}^{(1)\trsp} \bm{X}\bm{w}^{(2)} + b ) \nonumber\\
&= f\big(\langle \; \bm{X}, \bm{w}^{(1)} \circ \bm{w}^{(2)} \; \rangle + b \big),
\label{Eq:MGLRM}
\end{align}
where $\bm{w}^{(1)}$ and $\bm{w}^{(2)}$ are vectors of weights. 
Following a similar procedure, the model can be generalized to $M$-dimensional tensor-valued data with
\begin{equation}
y= f \big(\langle \; \bm{\mathcal{X}}, \bm{w}^{(1)} \circ \ldots \circ \bm{w}^{(M)} \; \rangle + b \big),
\label{Eq:TGLRM}
\end{equation}
as shown in Figure~\ref{Fig:GLRMmodels}-\emph{middle}.
The key advantages of this representation, compared to vector-valued representation $y = f(\text{vec}(\bm{\mathcal{X}})^\trsp\bm{w} + b)$, are that the underlying structure of the tensor-valued data is taken into account in the model and that the number of parameters is reduced from $\prod_{m=1}^{M}I_m$ to $\sum_{m=1}^{M}I_m$.
Moreover, more complex features can be represented by encoding the weight tensor as a sum of $R$ rank-one tensors with
\begin{equation}
y= f \left(\Big\langle \; \bm{\mathcal{X}}, \sum_{r=1}^{R} \bm{w}_r^{(1)} \circ \ldots \circ \bm{w}_r^{(M)} \; \Big\rangle + b \right).
\label{Eq:TGLRMrank}
\end{equation}
This model is represented in Figure~\ref{Fig:GLRMmodels}-\emph{bottom}.

Similarly to the vector case, we obtain the tensor-valued linear and logistic regression models by defining the function $f(\cdot)$ as identity and as the softmax function, respectively.

\subsection{Tensor ridge regression (TRR)}
Given a vector-valued input data $\bm{x}$, the classical regression model is of the form
\begin{equation}
y = \bm{x}^\trsp\bm{w} + b + \epsilon = \langle \bm{x}, \bm{w}\rangle +b + \epsilon,
\label{Eq:RR}
\end{equation}
where $y$ is the predicted output, $\bm{w}$ is a vector of weights, $b$ is the bias and $\epsilon$ is a zero-mean Gaussian noise variable.
Following \eqref{Eq:TGLRMrank}, and as shown in \cite{Guo12,Zhou13}, the model can be generalized to $M$-dimensional tensor-valued data 
\begin{align}
y&= \langle \; \bm{\mathcal{X}}, \sum_{r=1}^{R} \bm{w}_r^{(1)} \circ \ldots \circ \bm{w}_r^{(M)} \; \rangle + b + \epsilon \nonumber \\
&= \langle \; \bm{\mathcal{X}}, \bm{\mathcal{W}} \; \rangle + b + \epsilon,
\label{Eq:TRRrank}
\end{align}
therefore taking the underlying structure of the data into account.

Given a dataset of $M$-dimensional tensor inputs and corresponding outputs $\{\bm{\mathcal{X}}_n,y_n\}_{n=1}^N$, the parameters of the tensor ridge regression (TRR) model \eqref{Eq:TRRrank} are learned by maximizing its likelihood, or equivalently its corresponding log-likelihood
\begin{equation}
\ell(\bm{\mathcal{W}}, b, \sigma) = \sum_{n=1}^{N} \log \mathcal{N} \Big( y_n\big|\langle \; \bm{\mathcal{X}}_n, \bm{\mathcal{W}} \; \rangle + b, \, \sigma^2 \Big),
\label{Eq:TRRloglikelihood}
\end{equation}
where $\sigma^2$ is the variance of the zero-mean Gaussian random variable $\epsilon$. By using the inner product equality \eqref{Eq:InnerProdEquivalence2}, one can observe that the model \eqref{Eq:TRRrank} is linear in $\bm{W}^{(m)} =[\bm{w}_1^{(m)} \ldots \bm{w}_R^{(m)}]$ individually, so that the parameters \\
$\{\bm{W}^{(1)},\ldots,\bm{W}^{(M)},b\}$ can be learned by optimizing a sequence of generalized linear models (see \cite{Guo12,Zhou13} for details). 
Therefore, by adding to the log-likelihood function a zero-mean Gaussian prior on the weight tensor, equivalent to the regularization term $-\lambda_{\bm{\mathcal{W}}} \sum_{m=1}^{M}\|\bm{W}^{(m)}\|_{\text{F}}^2$, the bias $b$ and factor matrices $\bm{W}^{(m)}$ are updated at each iteration until convergence with
\begin{align}
\text{vec}(\bm{W}^{(m)}) &\quad\gets\quad (\bm{\Phi}^\trsp \bm{\Phi} + \lambda_{\bm{\mathcal{W}}} \bm{I})^{-1}\bm{\Phi}^\trsp (\bm{y}-b\,\bm{1}), 
\label{Eq:TRRUpdateW} \\
b &\quad\gets\quad \frac{1}{N} \sum_{n=1}^{N} y_n - \langle \bm{\mathcal{X}}_n, \bm{\mathcal{W}}\rangle,
\label{Eq:TRRUpdateB}
\end{align}
where the $n$-th row of the matrix $\bm{\Phi}$ is equal to \\
$\text{vec}\big( \bm{X}_{n,(m)}\bm{W}^{(-m)} \big)$, the $n$-th element of the vector $\bm{y}$ is $y_{n}$, $\bm{1}\in\mathbb{R}^N$ is a vector of $N$ ones and $\|\cdot\|_{\text{F}}$ is the Frobenius norm.
Note that other types of regularization are also proposed in \cite{Guo12}.

\subsection{Tensor logistic regression (TLR)}
In the classical multi-class logistic regression model, the posterior probability of the class $\mathcal{C}_i$ is given by the softmax function 
\begin{equation}
p(\mathcal{C}_i|\bm{x}, \bm{\theta}) = \frac{\exp(\bm{x}^\trsp \bm{v}_i+a_i)}{\sum_{j=1}^{C}\exp(\bm{x}^\trsp \bm{v}_j+a_j)},
\label{Eq:LRsoftmax}
\end{equation}
where $\bm{\theta}$ denotes the parameters of the model and $C$ the number of classes.
Similarly as ridge regression, the logistic regression model can be extended to tensor-valued data by encoding the tensor of weights as a sum of $R$ rank-one tensors, leading to the tensor-valued softmax function \cite{Hung13,Tan13}
\begin{equation}
\pi_i = p(\mathcal{C}_i|\bm{\mathcal{X}}, \bm{\theta}) = \frac{\exp \Big( \langle \bm{\mathcal{X}}, \bm{\mathcal{V}}_i \rangle + a_i \Big)}{\sum_{j=1}^{C}\exp \Big( \langle \bm{\mathcal{X}}, \bm{\mathcal{V}}_j \rangle +a_j \Big) },
\label{Eq:TLRsoftmax}
\end{equation}
where $\bm{\mathcal{V}}_i = \sum_{r=1}^{R} \bm{v}_{i,r}^{(1)} \circ \ldots \circ \bm{v}_{i,r}^{(M)}$.
Similarly to TRR, the tensor logistic regression (TLR) model takes into account the underlying structure of the data and reduces the number of parameters in the model compared to a vector-based representation of the tensor-valued data.

Given a dataset of inputs and corresponding unit vector outputs $\{\bm{\mathcal{X}}_n,y_n\}_{n=1}^N$, where $y_{n,i}=1$ indicates that the $n$-th data belong to the $i$-th class, the log-likelihood of the multivariate tensor logistic regression model is
\begin{align}
&\ell\big(\{\bm{\mathcal{V}}_i,a_i\}_{i=1}^{C}\big) = \log \prod_{n=1}^{N}\prod_{i=1}^{C} \pi_i^{y_{n,i}} \nonumber \\
&= \begin{small}
\sum_{n=1}^{N} \bigg(\! \sum_{i=1}^{C} y_{n,i} \big(\langle \bm{\mathcal{X}}_n, \bm{\mathcal{V}}_i \rangle \!+\! a_i \big) - \log \sum_{i=1}^{C} \exp\!\big(\langle \bm{\mathcal{X}}_n, \bm{\mathcal{V}}_i \rangle \!+\! a_i\big)\! \bigg).
\end{small} 
\label{Eq:TLRloglikelihood}
\end{align}
Note that a regularization term $-\lambda_{\bm{\mathcal{V}}} \sum_{i=1}^{C}\sum_{m=1}^{M}\|\bm{V}_i^{(m)}\|_{\text{F}}^2$ can be added to the log-likelihood function to avoid overfitting.
The parameters $\{\bm{V}_i^{(1)},\ldots,\bm{V}_i^{(M)},a_i\}_{i=1}^{C}$ can be learned by minimizing the negative log-likelihood of the model via any gradient-based optimizer, e.g., Newton's method or limited memory BFGS. By exploiting \eqref{Eq:InnerProdEquivalence2}, the gradients of the regularized negative log-likelihood used in the optimization process can be computed as
\begin{align}
\frac{\delta\big(-\ell(\{\bm{\mathcal{V}}_i,a_i\}_{i=1}^{C})\big)}{\delta \bm{V}_i^{(m)}} = &\sum_{n=1}^{N} (\pi_{n,i}-y_{n,i}) \text{vec}(\bm{X}_{n,(m)}\bm{V}_i^{(-m)}) \nonumber 
\label{Eq:TLRllGrad_V} \\
& + 2\,\lambda_{\bm{\mathcal{V}}}\, \text{vec}(\bm{V}_i^{(m)}), \\
\frac{\delta\big(-\ell(\{\bm{\mathcal{V}}_i,a_i\}_{i=1}^{C})\big)}{\delta a_i} = &\sum_{n=1}^{N} (\pi_{n,i}-y_{n,i}),
\label{Eq:TLRllGrad_a}
\end{align}
where $\pi_{n,i} = p(\mathcal{C}_i|\bm{\mathcal{X}}_n, \theta)$.

\section{Tensor-variate mixture of experts}
\label{sec:TensorMixtExp}
A mixture-of-experts (ME) regression model \cite{Jacobs91} aims at solving a nonlinear supervised learning problem by combining the predictions of a set of experts. The model is composed of a gate determining a soft division of the input space, and several experts making predictions in the different regions of the input space. In this section, we propose to generalize the ME regression model to tensor-variate data by using tensor-variate models for the experts and for the gate.

\subsection{TME model}
Given a tensor-variate input $\bm{\mathcal{X}}$ and an output $\bm{y}$, the tensor-variate mixture distribution is
\begin{equation}
p(\bm{y}|\bm{\mathcal{X}}, \bm{\theta}) = \sum_{i=1}^{C} \, p(i|\bm{\mathcal{X}}, \bm{\theta}_g) \; p(\bm{y}|i, \bm{\mathcal{X}},\bm{\theta}_e),
\label{Eq:TME}
\end{equation} 
where $C$ is the number of experts, $p(i|\bm{\mathcal{X}},\bm{\theta}_g)$ is the probability of the $i$-th expert to be activated (gate's rating) and $p(\bm{y}|i, \bm{\mathcal{X}},\bm{\theta}_e)$ is the model of the $i$-th expert. We define $\bm{\theta}=\{\bm{\theta}_g, \bm{\theta}_e\}$, where $\bm{\theta}_g$ and $\bm{\theta}_e$ denote the parameters of the gate and the set of experts, respectively.

Similarly to the original ME model, we define the gate of the TME model by the tensor-variate softmax function, so that
\begin{equation}
p(i|\bm{\mathcal{X}},\bm{\theta}_g)=\pi_i,
\label{Eq:TMEgateModel}
\end{equation}
with $\pi_i$ defined by \eqref{Eq:TLRsoftmax}, $\bm{\mathcal{V}}_i = \sum_{r=1}^{Rg} \bm{v}_{i,r}^{(1)} \circ \ldots \circ \bm{v}_{i,r}^{(M)}$ and $R_{g}$ the rank of the weight tensors $\bm{\mathcal{V}}_i$.
The experts follow the Gaussian model
\begin{equation}
p(\bm{y}|i,\bm{\mathcal{X}}, \bm{\theta}_e) = \mathcal{N} \Big(\bm{y} \big| \bm{\psi}_i(\bm{\mathcal{X}}) + \bm{b}_{i}, \bm{\Sigma}_{i} \Big),
\label{Eq:TMEexpModel}
\end{equation}
where $\bm{\psi}_i(\bm{\mathcal{X}}) =
\left(\begin{smallmatrix}
\langle\; \bm{\mathcal{X}}, \bm{\mathcal{W}}_{i,1} \;\rangle \\
\vdots \\
\langle\; \bm{\mathcal{X}}, \bm{\mathcal{W}}_{i,D} \;\rangle
\end{smallmatrix} \right)$, $\bm{\mathcal{W}}_{i,d} = \sum_{r=1}^{R_{e_i}} \bm{w}_{i,d,r}^{(1)} \circ \ldots \circ \bm{w}_{i,d,r}^{(M)}$ and $R_{e_i}$ is the rank of the weight tensors $\bm{\mathcal{W}}_{i,d}$. Note that one weight tensor $\bm{\mathcal{W}}_{i,d}$ is defined for each element of $\bm{y}\in\mathbb{R}^D$. This is similar to the vector case, where different vectors $\bm{w}_i$ weight the input for each element of the output, so that $\bm{y}=\left(\begin{smallmatrix}
\bm{w}_1^\trsp \\ \vdots \\ \bm{w}_D^\trsp \\
\end{smallmatrix}\right)\bm{x} + \bm{b}$.

\begin{figure}[tbp]
	\centering
	\includegraphics[width=0.48\textwidth]{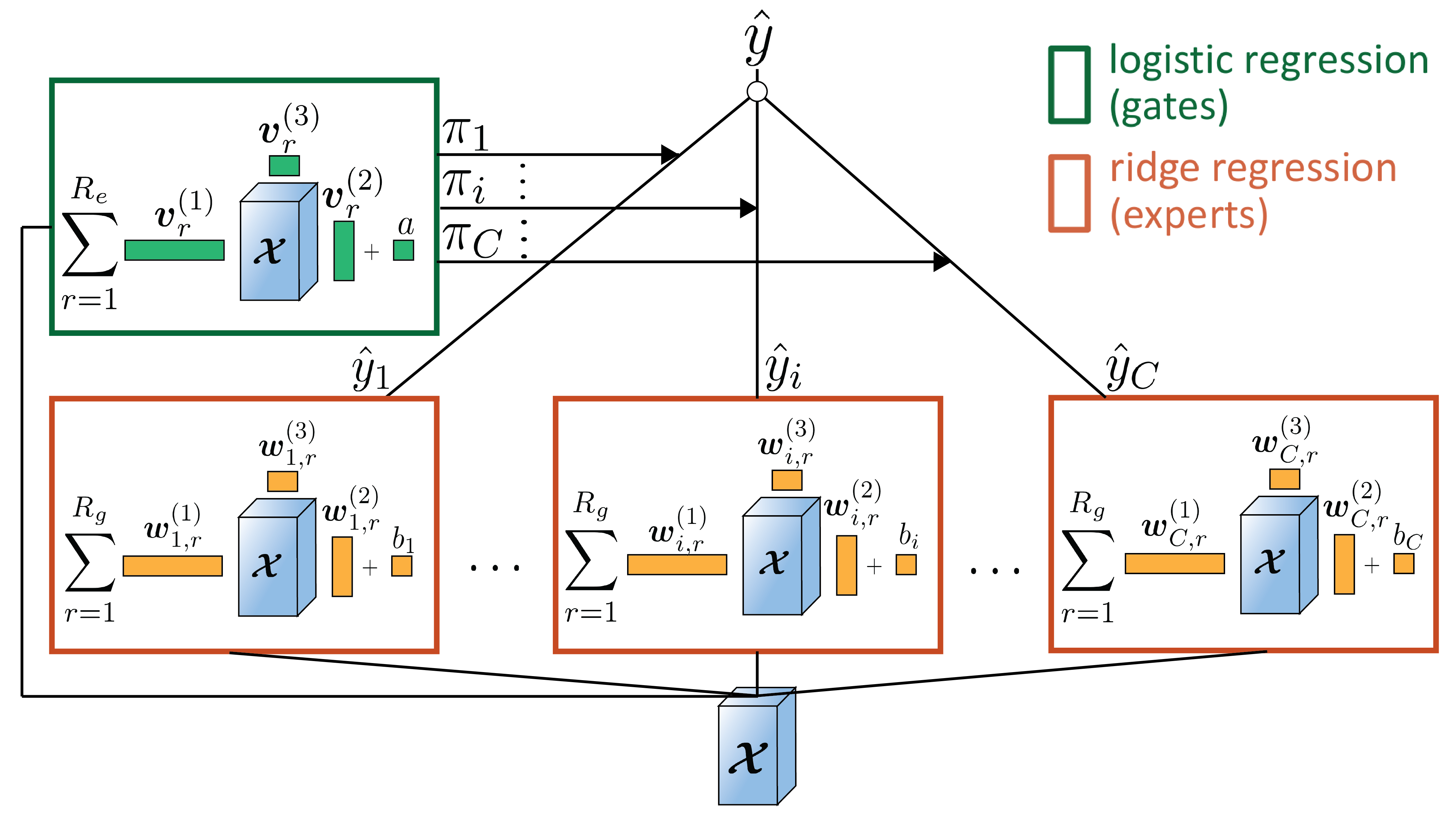}
	\caption{Proposed TME model. The gate is represented in the green box and the experts are represented in orange boxes.}  
	\label{Fig:TMEmodel}
\end{figure}

Figure~\ref{Fig:TMEmodel} illustrates the proposed model. Single predictions are computed by using the expectation of the TME model \eqref{Eq:TME}, so that
\begin{equation}
\bm{\hat{y}} = \sum_{i=1}^{C} 
\pi_i
(\bm{\psi}_i(\bm{\mathcal{X}}) + \bm{b}_{i}).
\label{Eq:TMEprediction}
\end{equation}

\subsection{Training of TME}
Similarly to ME, the TME model can be trained using the expectation-maximization (EM) algorithm. By introducing a set of binary latent variables $\{\bm{z}_n\}$ where $z_{n,i}=1$ indicate that the data $n$ was generated by the $i$-th mixture component, the expected complete data log-likelihood is given by
\begin{equation}
Q(\bm{\theta}) = \sum_{n=1}^{N} \sum_{i=1}^{C} r_{n,i} \log\Big( \pi_{n,i} \mathcal{N}\big(\bm{y}_n \big| \bm{\psi}_i(\bm{\mathcal{X}}_n) + \bm{b}_{i}, \bm{\Sigma}_{i} \big) \Big),
\label{Eq:TMEcompleteLL}
\end{equation}
where $r_{n,i}$ denotes the responsibility of the $i$-th component for the $n$-th data point so that $r_{n,i} = p(z_{n,i}=1|\bm{\mathcal{X}}_n,\bm{\theta})$.
In the E-step, the responsibilities $r_{n,i}$ are computed using
\begin{equation}
r_{n,i}=
\frac{\pi_{n,i} \; \mathcal{N}\big(\bm{y}_n \big| \bm{\psi}_i(\bm{\mathcal{X}}_n) + \bm{b}_{i}, \bm{\Sigma}_{i} \big)} 
{\sum_{j=1}^{C} \pi_{n,j} \; \mathcal{N}\big(\bm{y}_n \big| \bm{\psi}_j(\bm{\mathcal{X}}_n) + \bm{b}_{j}, \bm{\Sigma}_{j} \big)}.
\label{Eq:TMEeStep}
\end{equation}

In the M-step, the parameters are updated to maximize the expected complete data log-likelihood \eqref{Eq:TMEcompleteLL}. First, the parameters of the experts $\bm{\theta}_e$ are updated iteratively until convergence, based on \eqref{Eq:TRRUpdateW} and \eqref{Eq:TRRUpdateB}, with
\begin{align}
\text{vec}(\bm{W}_{i,d}^{(m)}) &\gets (\bm{\tilde{\Phi}}_{i,d}^\trsp \bm{\tilde{\Phi}}_{i,d} + \lambda_{\bm{\mathcal{W}}} \bm{I})^{-1}\bm{\tilde{\Phi}}_{i,d}^\trsp (\bm{\tilde{y}}_d - b_{i,d}\bm{1}), \\
b &\gets \frac{1}{N} \sum_{n=1}^{N} \tilde{y}_{n,d} - \langle \bm{\tilde{\mathcal{X}}}_n, \bm{\mathcal{W}}_{i,d}\rangle,
\label{Eq:TMEmstepTRR}
\end{align}
where the $n$-th row of the matrix $\bm{\tilde{\Phi}}_{i,d}$ is equal to \\
$\text{vec}\big( \bm{\tilde{X}}_{n,(m)}\bm{W}_{i,d}^{(-m)} \big)$, the $n$-th element of the vector $\bm{\tilde{y}}$ is $\tilde{y}_{n,d}$ and
$\bm{\tilde{\mathcal{X}}}_n = \sqrt{r_{nk}}\bm{\mathcal{X}}_n$, $\bm{\tilde{y}}_n = \sqrt{r_{nk}}\bm{y}_n$ are the scaled input tensors and output vectors, respectively. The covariance of the experts Gaussian model is then updated as
\begin{equation}
\bm{\Sigma}_i \gets \frac{\sum_{n=1}^{N}r_{n,i} \big(\bm{y}_n-\bm{\psi}_i(\bm{\mathcal{X}}_n)-\bm{b}_i \big)^{\!\trsp} \big(\bm{y}_n-\bm{\psi}_i(\bm{\mathcal{X}}_n)-\bm{b}_i \big) }
{\sum_{n=1}^{N}r_{n,i}}.
\label{Eq:TMEmstepSigma}
\end{equation}

Finally, the gate parameters $\bm{\theta}_g$ are updated by maximizing the log-likelihood of the multivariate tensor logistic regression model 
\begin{equation}
\ell\big(\{\bm{\mathcal{V}}_i,a_i\}_{i=1}^{C}\big) = \log \prod_{n=1}^{N}\prod_{i=1}^{C} \pi_{n,i}^{r_{n,i}} -\lambda_{\bm{\mathcal{V}}} \sum_{i=1}^{C}\sum_{m=1}^{M}\|\bm{V}_i^{(m)}\|_{\text{F}}^2,
\label{Eq:TMEmstepTLR}
\end{equation}
based on \eqref{Eq:TLRloglikelihood}. Similarly to tensor logistic regression, a gradient-based optimizer is used to minimize the negative log-likelihood with gradients given by \eqref{Eq:TLRllGrad_V} and \eqref{Eq:TLRllGrad_a}, where $y_{n,i}$ is replaced by $r_{n,i}$. 

Note that regularization terms have been added in the M-step to avoid overfitting. The E-step and M-step are iterated until convergence of the algorithm.

\subsection{Model selection and initialization}
Selecting the number of experts in ME is known to be a difficult problem \cite{Yuksel12}. When the structure of the application allows it, as in the experiments of Section~\ref{sec:Experiments}, the number of experts can be determined by the experimenter. Otherwise, standard strategies used for ME, such as exhaustive search, growing or pruning models, or Bayesian estimates can be adapted to TME.

The TME model assumes fixed ranks $R_g$ and $R_{e_i}$ for the gate and experts weight tensors, respectively. The appropriate rank can be estimated using cross-validation or through usual model selection criterion, e.g., the Bayesian information criterion (BIC). 

Previous works on TRR \cite{Guo12,Zhou13} and TLR \cite{Tan13} showed that both TRR and TLR models converge to a similar solution independently of the initial weight values. Therefore, the weight tensors of TME are initialized with random values in our experiments. In order to facilitate the convergence, we initialized the weights of the expert model $\bm{\mathcal{W}}_i$ and of the gate $\bm{\mathcal{V}}_i$ as equal to the weights $\bm{\mathcal{W}}$ obtained from TRR.

\section{Experiments}
\label{sec:Experiments}
In this section, we first evaluate the functionality and the performance of the proposed TME on artificially generated data. The approach is then applied to the detection and recognition of hand movements from tactile myography (TMG) data. An offline experiment and a real-time teleoperation experiment, where participants controlled a robotic arm and hand based on their hand movements, illustrate the effectiveness of the proposed TME model. A video of the teleoperation experiment accompanies the paper (\url{https://sites.google.com/view/tensor-mixture-of-experts/}). Source codes related to the experiments are available at \url{https://github.com/NoemieJaquier/TME}. 

\subsection{2D shape example}
\label{subsec:Experiment2d}
\begin{figure}[tbp]
	\centering
	\begin{subfigure}[t]{0.07\textwidth}
		\captionsetup{justification=centering}
		\includegraphics[width=\textwidth]{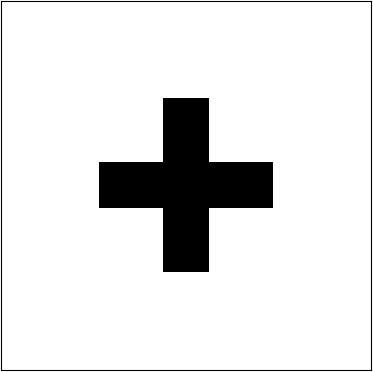}
		\includegraphics[width=\textwidth]{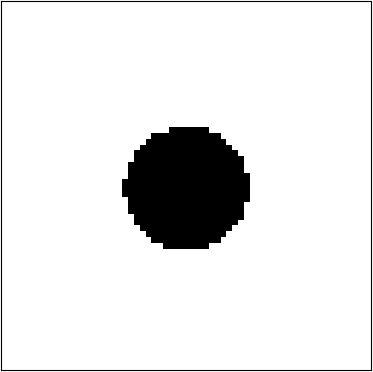}
		\includegraphics[width=\textwidth]{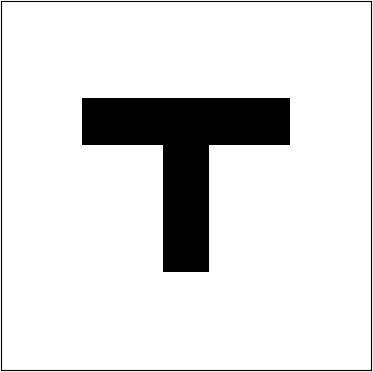}
		\caption{} 
		\label{subFig:Dummy01Original}
	\end{subfigure}
	\begin{subfigure}[t]{0.07\textwidth}
		\captionsetup{justification=centering}
		\includegraphics[width=\textwidth]{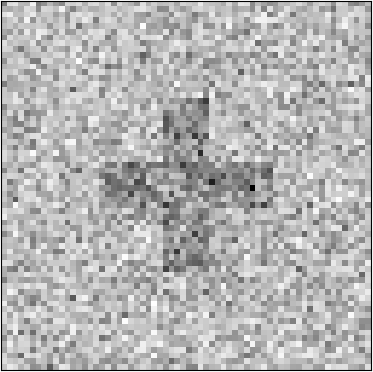}
		\includegraphics[width=\textwidth]{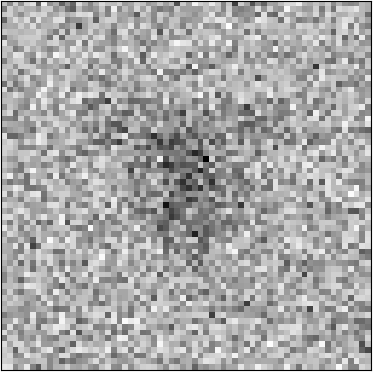}
		\includegraphics[width=\textwidth]{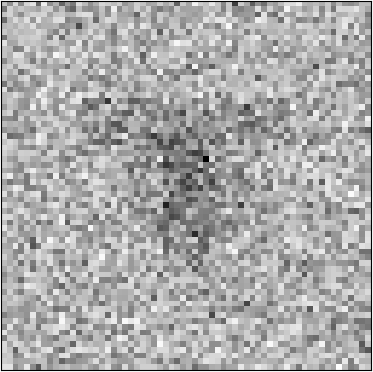}
		\caption{}
		\label{subFig:Dummy01ME}
	\end{subfigure}	
	\begin{subfigure}[t]{0.07\textwidth}
		\captionsetup{justification=centering}
		\includegraphics[width=\textwidth]{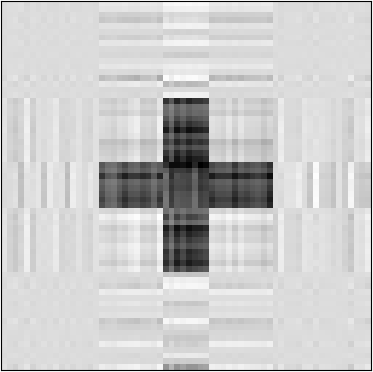}
		\includegraphics[width=\textwidth]{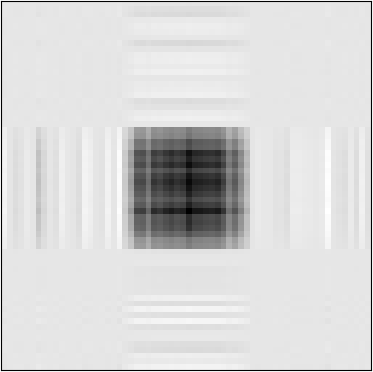}
		\includegraphics[width=\textwidth]{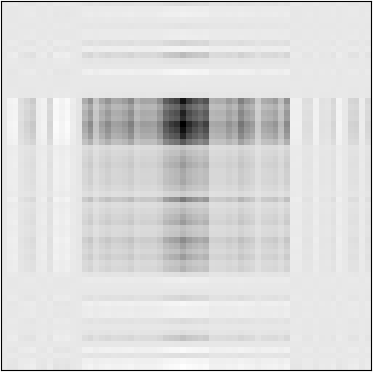}
		\caption{} 
		\label{subFig:Dummy01MME_Rank-RR11-LR2}
	\end{subfigure}
	\begin{subfigure}[t]{0.07\textwidth}
		\captionsetup{justification=centering}
		\includegraphics[width=\textwidth]{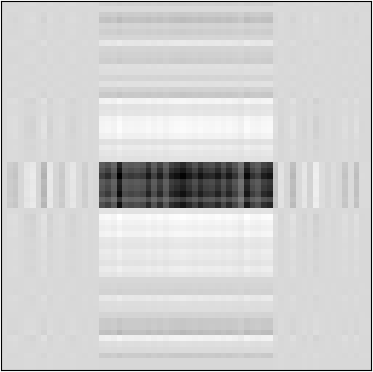}
		\includegraphics[width=\textwidth]{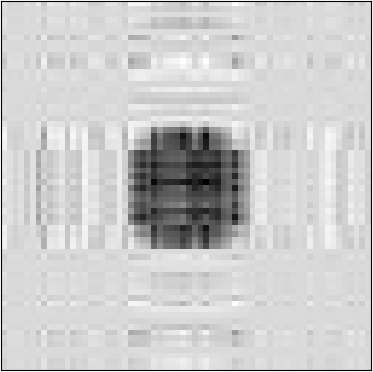}
		\includegraphics[width=\textwidth]{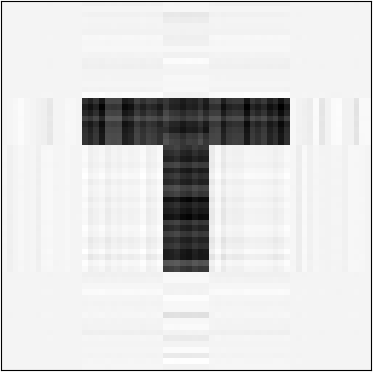}
		\caption{} 
		\label{subFig:Dummy01MME_Rank-RR22-LR1}
	\end{subfigure}	
	\begin{subfigure}[t]{0.07\textwidth}
		\captionsetup{justification=centering}
		\includegraphics[width=\textwidth]{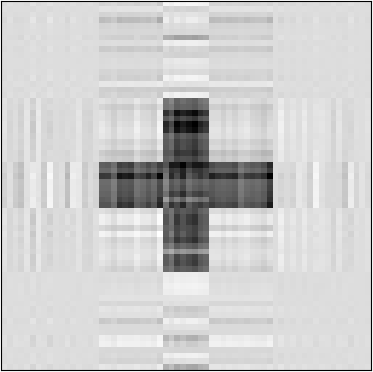}
		\includegraphics[width=\textwidth]{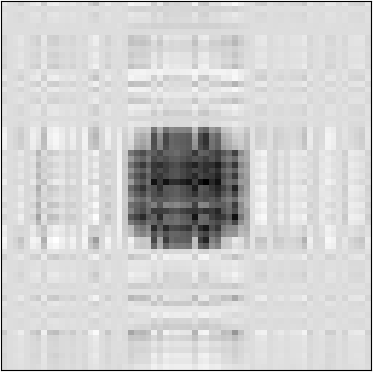}
		\includegraphics[width=\textwidth]{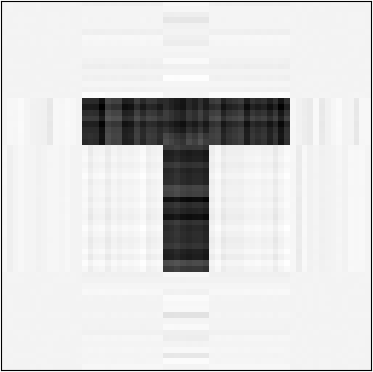}
		\caption{} 
		\label{subFig:Dummy01MME_Rank-RR22-LR2}
	\end{subfigure}	
	\begin{subfigure}[t]{0.07\textwidth}
		\captionsetup{justification=centering}
		\includegraphics[width=\textwidth]{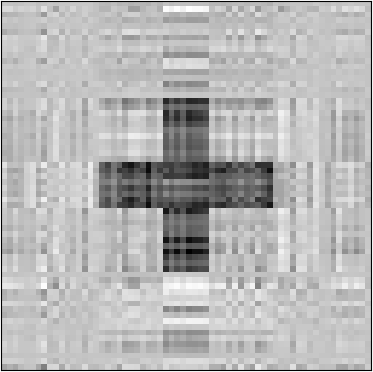}
		\includegraphics[width=\textwidth]{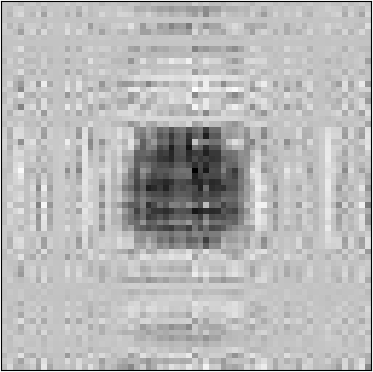}
		\includegraphics[width=\textwidth]{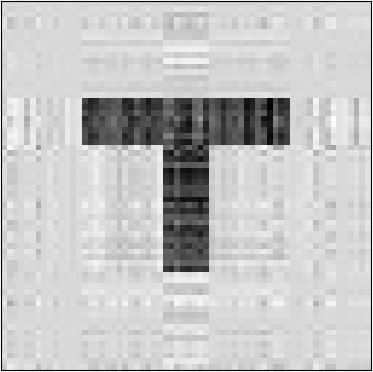}
		\caption{} 
		\label{subFig:Dummy01MME_Rank-RR33-LR3}
	\end{subfigure}	
	\begin{minipage}[b]{0.45\textwidth}
		\renewcommand*{\arraystretch}{1.2}
		\small
		\begin{tabular*}{\textwidth}{c @{\extracolsep{\fill}} ccccc}
			& ME & TME & TME & TME & TME  \\
			& &(2,1) & (1,2) & (2,2) & (3,3) \\
			\hline
			RMSE $\;\;$ & $0.30$ & $0.16$ & $0.18$ & $0.12$ & $0.18$\\
			\hline
			BIC & - & $10209$ & $8940$ & $9835$ & $11216$ \\
			\hline
		\end{tabular*}
	\end{minipage}
	\caption{(\emph{a}) True weight matrices of the gate and experts functions (from \emph{top} to \emph{bottom}: $\bm{V}$, $\bm{W}_{1}$ and $\bm{W}_{2}$). (\emph{b}) Recovered weight matrices by ME. (\emph{c-f}) Recovered weight matrices by TME for different ranks $(R_g, R_e)$.}  
	\label{Fig:Dummy01Rank}
\end{figure}

In this illustrative example, we propose to evaluate the performance of the proposed TME model for different ranks under various sample sizes and signal strengths. To do so, we generate artificial data following the model \eqref{Eq:TME} from known parameters $\bm{\theta}$ and we evaluate the recovery of these parameters by the model. In this illustrative example, we consider matrix-variate inputs $\bm{X}\in\mathbb{R}^{64\times64}$ whose elements are independent and normally distributed. The output $y$ is normally distributed with a mean given by a 2-classes TME model with zero biases
\begin{multline}
\hat{y} = \frac{\exp\big(\langle \bm{X}, \bm{V} \rangle\big)}{1 + \exp\big(\langle \bm{X}, \bm{V} \rangle\big)} \; \langle \bm{X}, \bm{W}_1 \rangle \;+\\ 
\frac{1}{1 + \exp\big(\langle \bm{X}, \bm{V} \rangle\big)} \; \langle \bm{X}, \bm{W}_2 \rangle,
\label{Eq:DummyMeanModel}
\end{multline}
and a standard deviation $\sigma$. The weight matrices $\bm{V}, \bm{W}_{1}, \bm{W}_{2} \in\mathbb{R}^{64\times64}$ are equal to the binary 2D shapes represented in Figure~\ref{subFig:Dummy01Original}, where the black and white regions correspond to $1$ and $0$, respectively. The use of these 2D shapes was inspired by the illustrative example presented in \cite{Zhou13}.

We first examine the performances of the proposed TME model for ranks $R_g$ and $R_e = R_{e_1} = R_{e_2}$ varying from $1$ to $3$ with a sample size $N=1000$ equally divided between the two classes and a noise level $\sigma$ equal to $10\%$ of the standard deviation of the mean $\hat{y}$. The regularization terms $\lambda_{\bm{V}}$ and $\lambda_{\bm{\mathcal{W}}}$ were fixed as equal to $0.1$. Moreover, we compare the TME model with the standard ME regression model whose gate is defined by the softmax function \eqref{Eq:LRsoftmax} and experts follow a Gaussian model with a mean given by \eqref{Eq:RR}.

Figure~\ref{Fig:Dummy01Rank} shows the original and recovered weight matrices by the ME and TME models along with the root-mean-square error (RMSE) for the estimation of the weight matrices and the BIC value for TME. We observe that TME outperforms ME for all the tested rank values as the maximum RMSE value achieved by TME is $0.21$ ($R_g=1$, $R_e=3$) versus $0.3$ for the ME model. Moreover, we observe that the weight matrices retrieved by ME are noisier and the shapes of the experts weights are not clearly delimited and tend to be fused together compared to those retrieved by TME. Similar results were obtained for different sample sizes and noise levels. 

\begin{figure}[tbp]
	\centering
	\begin{subfigure}[t]{0.07\textwidth}
		\captionsetup{justification=centering}
		\includegraphics[width=\textwidth]{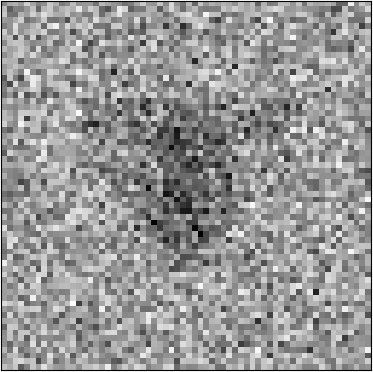}
		\caption{} 
		\label{subFig:Dummy01RR}
	\end{subfigure}	
	\begin{subfigure}[t]{0.07\textwidth}
		\captionsetup{justification=centering}
		\includegraphics[width=\textwidth]{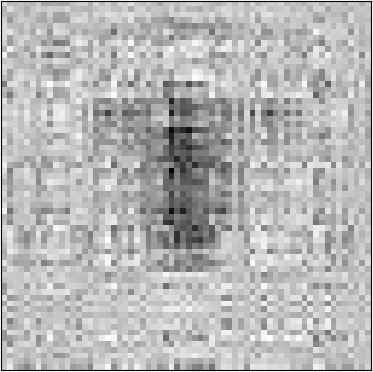}
		\caption{} 
		\label{subFig:Dummy01MMR}
	\end{subfigure}	
	\caption{(\emph{a}) Recovered weight matrix by RR. (\emph{b}) Recovered weight matrix by MRR.}  
	\label{Fig:Dummy01RR}
\end{figure}

Similar observations can also be made by comparing the weight matrices retrieved by RR and MRR for the same data, as shown in Figure~\ref{Fig:Dummy01RR}. Although both methods retrieve one weight matrix fusing the three original ones due to their formulation, the weight matrix retrieved by RR looks noisier than the one retrieved by MRR. This confirms that taking the structure of the data into account improves the quality of the recovered weight matrices.

Due to their structure, a rank-2 setting is sufficient to capture a cross or a t-shape pattern, while a low rank setting does not allow to exactly represent a disk shape. As expected, the cross and t-shape are fully recovered by TME in the cases where $R_g\geq2$ and $R_e\geq2$, respectively, while approximations of the shapes are obtained for lower ranks (see Figures~\ref{subFig:Dummy01MME_Rank-RR11-LR2}--\ref{subFig:Dummy01MME_Rank-RR33-LR3}). Moreover, while the disk shape is approximated by a square in a rank-1 setting (Figure~\ref{subFig:Dummy01MME_Rank-RR11-LR2}), it is already fairly recovered by a rank-2 or rank-3 setting (see Figures~\ref{subFig:Dummy01MME_Rank-RR22-LR1}--\ref{subFig:Dummy01MME_Rank-RR33-LR3}). 

Consistently with the aforementioned observations, the minimum RMSE value is obtained by TME with a rank-(2, 2) setting. Moreover, TME with ranks $R_g=R_e=3$ obtains a slightly higher RMSE than with ranks $R_g=R_e=2$. This can be explained by the fact that approximating the cross and t-shape with a rank-3 setting, while a rank-2 setting is sufficient, leads to an overfitted estimation with a higher influence of the noise contained in the training data. According to the BIC values reported for the tested TME models, the model with $R_g=1$ and $R_e=2$ should be selected (lowest BIC cost). However, in practice, one may prefer the rank-(2,2) setting in this case, suggesting that other rank selection methods, such as cross-validation, may be used in function of the application. Note that similar observations were made for different sample sizes and noise levels.

\begin{figure}[tbp]
	\centering
	\begin{subfigure}[b]{0.22\textwidth}
		\centering
		\includegraphics[width=\textwidth]{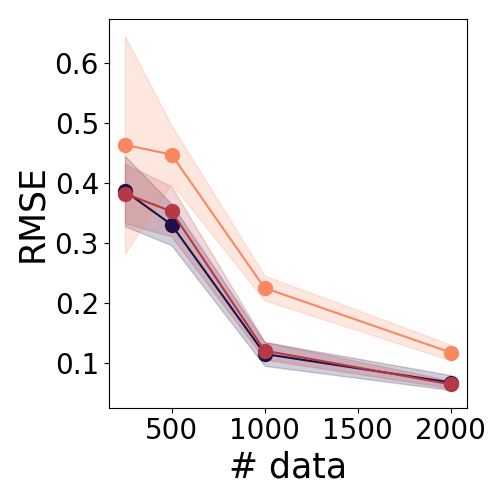}
		\caption{}
	\end{subfigure}
	\begin{subfigure}[b]{0.235\textwidth}
		\centering
		\includegraphics[width=0.31\textwidth]{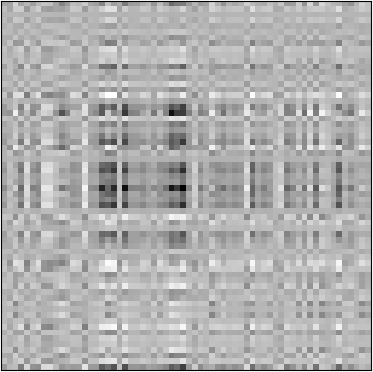}
		\includegraphics[width=0.31\textwidth]{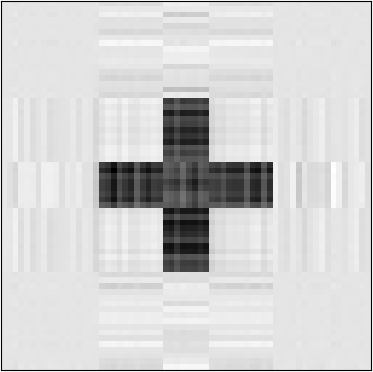}
		\includegraphics[width=0.31\textwidth]{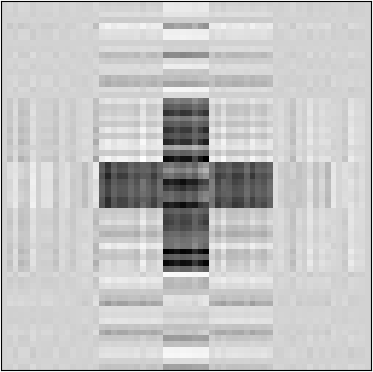}
		\includegraphics[width=0.31\textwidth]{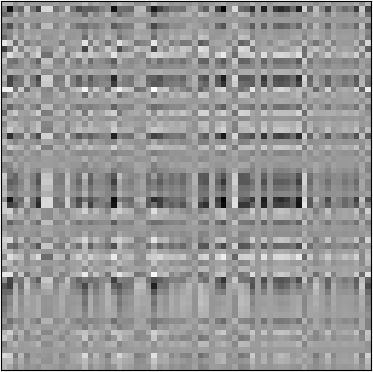}
		\includegraphics[width=0.31\textwidth]{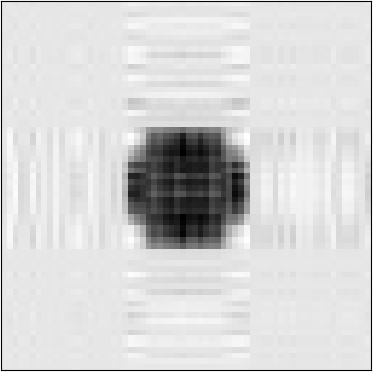}
		\includegraphics[width=0.31\textwidth]{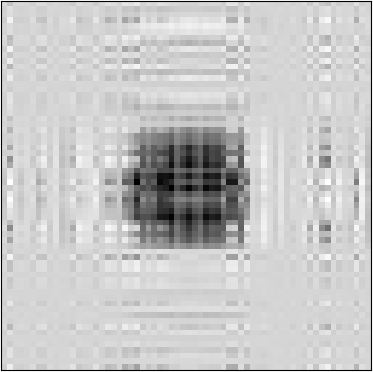}
		\includegraphics[width=0.31\textwidth]{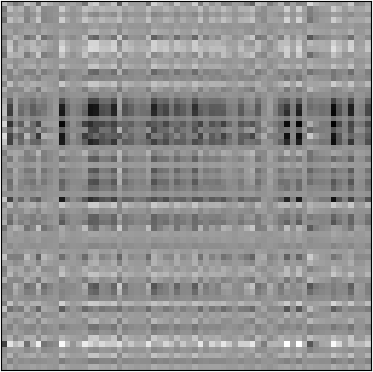}
		\includegraphics[width=0.31\textwidth]{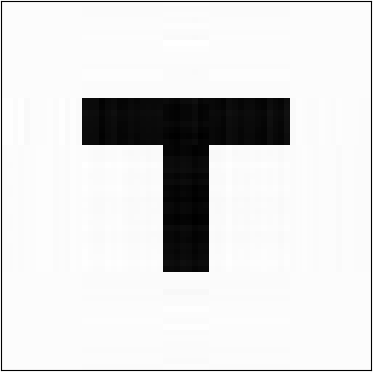}
		\includegraphics[width=0.31\textwidth]{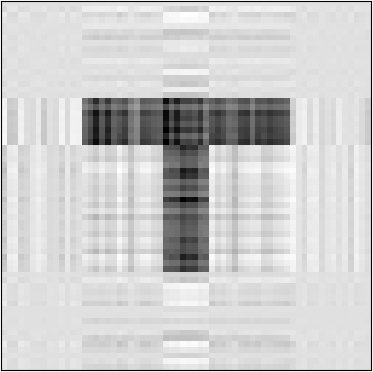}
		\caption{}
	\end{subfigure}
	\caption{(\emph{a}) Evolution of the RMSE of the estimation of TME weight matrices in function of the sample size $N$ for different noise levels. The curves corresponding to noise levels $\sigma$ equal to $1$, $10$ and $50\%$ of the standard deviation of the mean $\hat{y}$ are represented in dark blue, red and yellow, respectively. The mean and two standard deviations over 10 trials with different matrix-variate inputs $\bm{X}$ are represented. The sample size is equally divided between the two classes. (\emph{b}) Recovered weight matrices $\bm{V}$, $\bm{W}_{1}$ and $\bm{W}_{2}$ by TME for ranks $R_g=R_e=2$ with, from \emph{left} to \emph{right}, sample sizes $N=500$, $2000$ and $2000$ and noise levels $\sigma=1\%$, $1\%$ and $50\%$ of $\hat{y}$.}  
	\label{Fig:Dummy01Nbdata}
\end{figure}

As shown in Figure~\ref{Fig:Dummy01Nbdata}, the estimation accuracy increases with the sample size and decreases with the noise level $\sigma$, validating the consistency of the proposed method. 

\subsection{Shape example of higher dimensions}
\label{subsec:ExperimentHd}
In order to evaluate the performance of the proposed TME model for higher tensor dimensions, we extended the experiment to tensor-variate inputs $\mathcal{\bm{X}}$ of order 3 and 4. The dimension of the inputs and coefficients was reduced from $64$ to $16$ in order to allow the comparison with ME. Indeed, the number of elements of a third-order cube tensor of dimension $64$ is $262144$. Therefore, with a standard implementation of ME, $262144\times 262144$ matrices need to be stored and inverted, which cannot be handled in a straightforward manner with a standard computer. Note that other techniques, notably sparse methods, could be used to handle such a case. However, as this is out of the scope of this paper, we simply reduced the dimension of our coefficients.

For this second illustrative example, the tensor-variate inputs $\mathcal{\bm{X}}\in\mathbb{R}^{16\times \ldots\times 16}$ and the outputs $y$ were generated as in the previous experiment, with tensor coefficients instead of matrices in \eqref{Eq:DummyMeanModel}. The weight tensors $\mathcal{\bm{V}}$, $\mathcal{\bm{W}}_1$ and $\mathcal{\bm{W}}_2\in\mathbb{R}^{16\times \ldots\times 16}$ were defined as 2D, 3D and 4D binary coefficients with shapes similar to the ones represented in Figure~\ref{subFig:Dummy01Original} extended to higher dimensions. For this experiment, the background and shape regions correspond to 0 and 1. We compared the performances of ME with the proposed TME model with ranks $R_g = R_e=2$ with a sample size $N=200$ equally divided between the two classes and a noise level $\sigma$ equal to $10\%$ of the standard deviation of the mean $\hat{y}$. The regularization terms $\lambda_{\bm{V}}$ and $\lambda_{\bm{\mathcal{W}}}$ were fixed as equal to $0.1$. 

\begin{figure}[tbp]
	\centering
	\includegraphics[width=.25\textwidth]{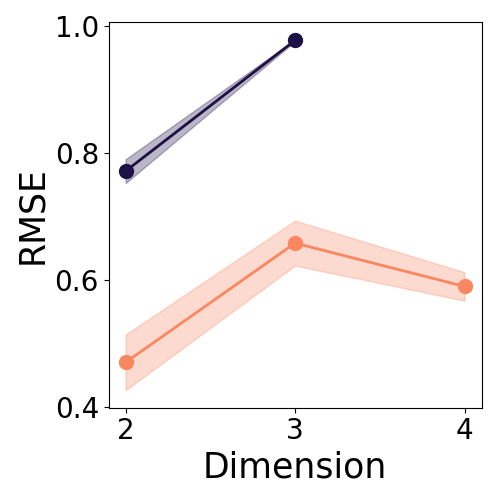}
	\caption{Evolution of the RMSE for the estimation of the weight tensors by ME and TME, depicted in dark blue and yellow, respectively. The mean and two standard deviations over 10 trials with different inputs are represented.}  
	\label{Fig:Dummy01dims}
\end{figure}

Figure~\ref{Fig:Dummy01dims} shows the mean and two standard deviations of the RMSE obtained for the estimation of the weight tensors by ME and TME for different dimensions. Note that no result is presented for ME with coefficients of dimension $4$ due to the computational load of storing and inverting $65536\times 65536$ matrices. We observe that TME outperforms ME for all dimensions. Moreover, the RMSE explodes for ME with dimension 3 as it almost reaches the maximum RMSE of 1, while it grows slowly for TME from dimensions 2 to 4. Considering that the number of training data was constant and that the number of elements of the coefficients grew considerably with the number of dimensions, this result highlights the benefits of TME (and of considering tensor-based approaches).

\subsection{Detection of hand movements from tactile myography}
\label{subsec:ExperimentTmg}
\begin{figure}[tbp]
	\centering
	\begin{subfigure}[b]{.75\columnwidth}
		\centering
		\includegraphics[width=\textwidth]{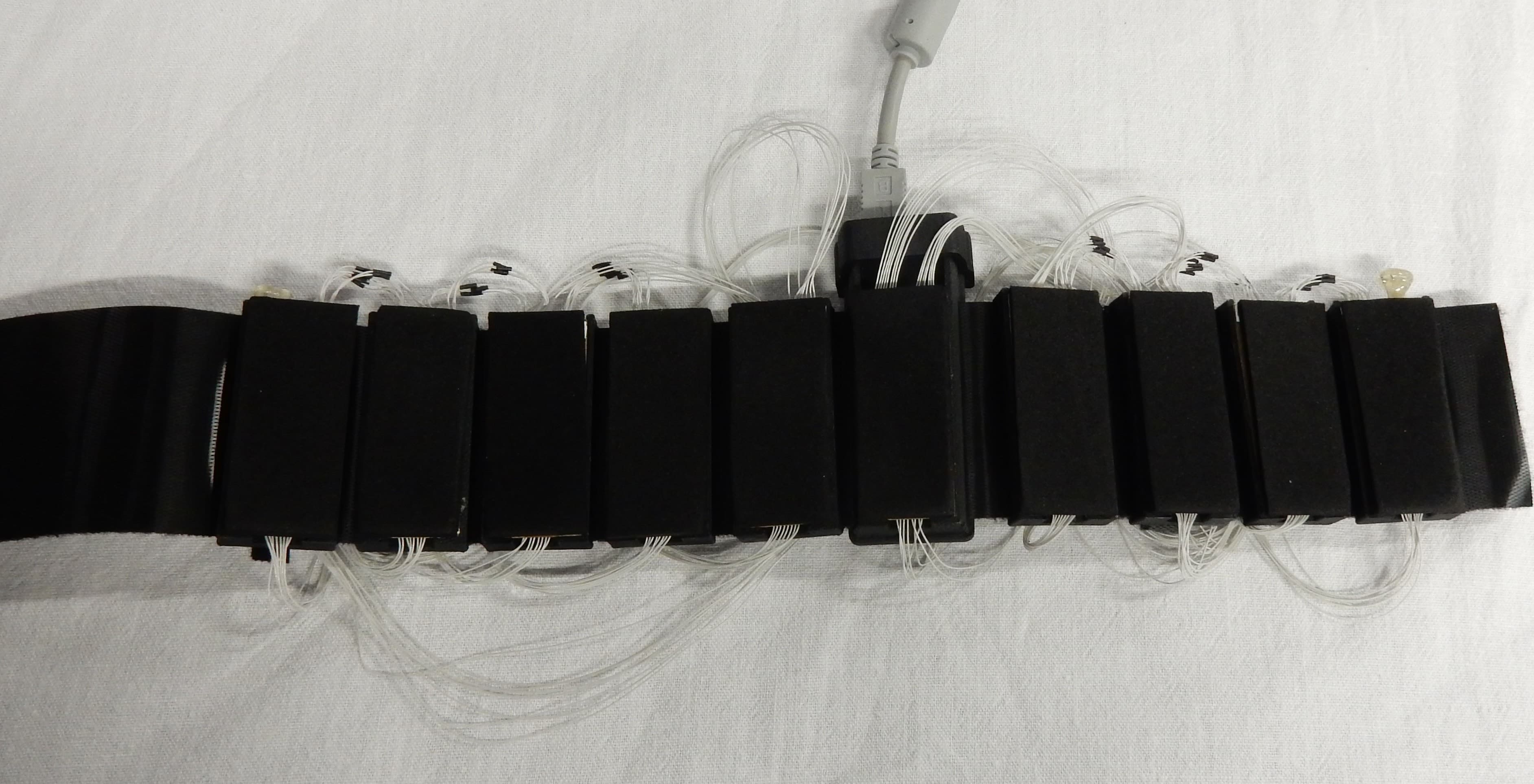}
		\caption{}
		\label{subFig:TactileBracelet}
	\end{subfigure}
	\\
	\begin{subfigure}[b]{.75\columnwidth}
		\centering
		\includegraphics[width=\textwidth]{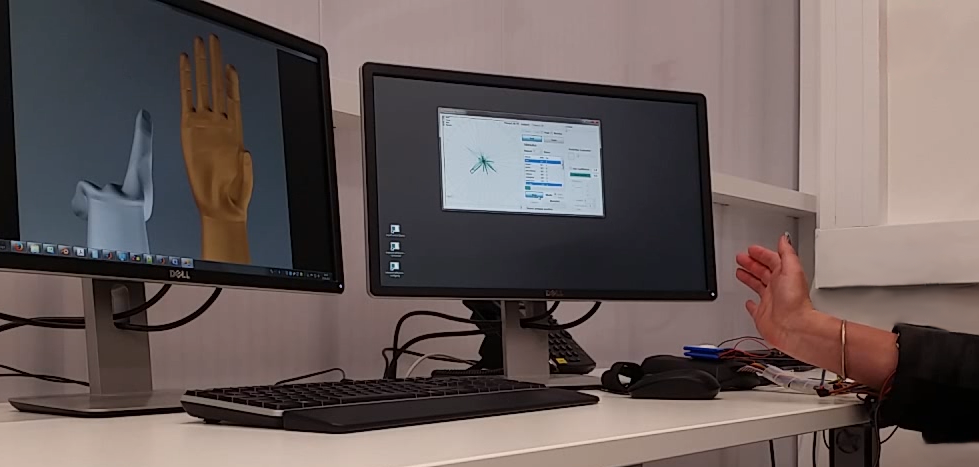}
		\caption{}
		\label{subFig:OfflineSetup}
	\end{subfigure}
	\caption{(\emph{a}) The TMG sensor used for the experiments. The bracelet is here rolled out, showing its $10$ modules of $8\times4$ resistive cells. (\emph{b}) Data collection \cite{Jaquier17}. The participant, wearing the tactile bracelet, imitates the gray animated hand model.}  
	\label{Fig:TMGsetup}
\end{figure}

In the context of prosthetic hands, tactile myography (TMG) has recently been proposed as a complementary or alternative approach to the traditional surface electromyography (sEMG) to achieve simultaneous and proportional control of multiples degrees of freedom (DOFs) of a hand prosthesis (see, e.g., \cite{Phillips05,Koiva15}). In this context, the aim of TMG is to measure the pressure related to the deformation induced by the muscles activity of the forearm. This signal is then used to determine the corresponding hand and wrist movements. Our TMG sensor, developed in \cite{Koiva15} and shown in Figure~\ref{subFig:TactileBracelet}, is composed of 320 resistive taxels organized in a $8\times40$ array forming a bracelet. Therefore, the data provided by the sensor are intrinsically matrix-valued.

Previous studies showed that ridge regression (RR) directly applied to the data of the bracelet allows the prediction of different finger and wrist movements \cite{Koiva15}, which could outperform detection using sEMG \cite{Nissler17}. However, RR does not take into account the matrix structure of the TMG data as they are vectorized before the application of the regression method. Moreover, the same weight vectors are used independently of the activated movements, which may result in false positive detection of activations.
Therefore, the motivations to use TME for this application are the following: (1) the structure of the data is taken into account in the regression process; (2) the problem is decomposed in two subparts, namely detecting which movements are activated and determining their individual level of activation; and (3) the low computational complexity to evaluate one test sample allows TME to be used for real-time detection of hand and wrist movements from TMG data.

In this experiment, we investigate the performance of TME on the dataset\footnote{The dataset is available online at \url{http://www.idiap.ch/paper/mdpi/data/exp2/}.} presented in the second experiment of \cite{Jaquier17}. The dataset was gathered from 9 healthy participants requested to replicate the movements of a stimulus in the form of a 9-DOF hand model while wearing the tactile bracelet. Ground truth was obtained from the values of the animated hand model displayed on a monitor. This method has the drawback of possibly reducing the precision of the prediction of the intended activations due to the delay required by the participant to replicate the displayed movement. However, this approach allows the association of intended activations with input signal patterns in the case of amputees (since ground truth data can obviously not be collected by other means in this case). 

Each participant executed three times a sequence of six movements, namely wrist flexion, wrist extension, wrist supination, thumb flexion, index flexion and little-finger flexion. The data were recorded during the whole cycle of the stimulus, namely transition, activation, transition and relaxation phases, in order to obtain the whole range of activation from rest to complete finger and wrist movements (see \cite{Jaquier17} for more details). 

The training dataset is composed of data recorded at zero and full activation. The testing dataset is composed of data recorded during the transition parts, containing the whole range of intermediate activation levels. Therefore, the evaluation of the performance of the model is compatible with the evaluation in forecasted studies with amputees, as they cannot provide accurate intermediate training data.

We compared the performance of vector-based and tensor-based algorithms on this dataset, namely RR, ME, TRR and TME. We also contrasted the results of these (mixtures of) linear models with a computationally more involved nonlinear method based on Gaussian process regression (GPR), see \cite{Jaquier17} for details.

The TMG data were centered for all the methods. The regularization parameter of RR and ME were fixed to $0.1$. GPR was used with a radial basis function (RBF) kernel whose parameters were optimized using GPy \cite{gpy14}. Note that GPR with RBF kernel with an Euclidean distance measure was proved to reach good performances on this dataset in \cite{Jaquier17}. The rank of TRR was determined using $5$-fold cross-validation and the regularization parameter $\lambda_{\bm{\mathcal{W}}}$ was fixed to $0.1$. For TME, as for ME, one expert was considered for each of the six finger and wrist movements. In order to facilitate the training process, we considered a common value $R_e$ for the ranks $R_{e_i}$. The ranks $R_g$ and $R_e$ were determined using $5$-fold cross-validation for $2\leq R_g, R_e \leq 6$. The regularization parameters of TME, $\lambda_{\bm{\mathcal{W}}}$ and $\lambda_{\bm{\mathcal{V}}}$, were fixed by the experimenter as equal to $0.1$. We noticed that small variations of these regularization parameters did not change significantly the results for the different regression methods. 

Table~\ref{Tab:TMGofflineMethodsRMSE} shows the mean and standard deviation over the 9 participants of the RMSE between the ground truth and the prediction for the aforementioned methods. 
We observe that taking into account the structure of the data in the regression process improves the quality of predictions as both TRR and TME outperform their vector counterparts. More surprisingly, taking into account the structure of the data allows a linear method (TRR) to achieve performance comparable to those obtained by a nonlinear method. 
GP and TME achieve the best performance compared to the other methods, with the linear TME approach obtaining only a slightly lower RMSE ($0.303\pm0.074$) than the nonlinear GP ($0.305\pm0.060$).\footnote{Note that GP with RBF kernel slightly outperformed GP with linear ($0.455\pm0.061$), Mat\'ern 32 ($0.308\pm0.061$) and Mat\'ern 52 ($0.306\pm0.059$) kernels, therefore all the results are presented with RBF kernel.} 
Moreover, TME obtained the minimum RMSE for 5 participants out of 9. 

\begin{figure}[tbp]
	\centering
	\includegraphics[width=0.235\textwidth]{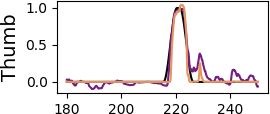}
	\vspace{0.2cm}
	\includegraphics[width=0.235\textwidth]{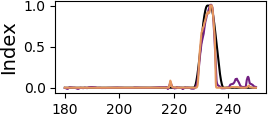}
	\vspace{0.2cm}
	\includegraphics[width=0.235\textwidth]{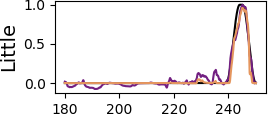}
	\includegraphics[width=0.235\textwidth]{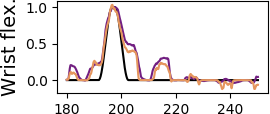}
	\includegraphics[width=0.235\textwidth]{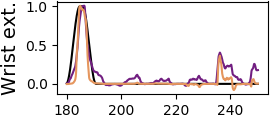}
	\includegraphics[width=0.235\textwidth]{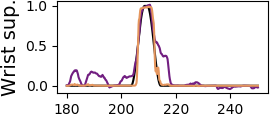}
	\includegraphics[width=0.22\textwidth]{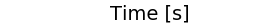}
	\includegraphics[width=0.22\textwidth]{S7_time}
	\caption{Original and recovered activations of the different fingers and wrist movements over time. The whole range of activation is represented from 0 to 1 on the vertical axis for each movement. The ground truth is shown by black curves, while the signals recovered by TME and GP are displayed in yellow and purple, respectively.}  
	\label{Fig:TMGresults}
\end{figure}

Figure~\ref{Fig:TMGresults} shows an example for GP and TME of the original and recovered activations for all movements over time. We observe that TME is generally recovering a more stable signal than GP when one movement is not activated. In the regions of zero activations, the signal recovered by GP tends to oscillate around zero. However, the signal recovered by TME can have a bigger delay than GP to detect an activation different than zero (see, e.g., Fig.~\ref{Fig:TMGresults}, wrist extension).

\begin{table}[t]
	\caption{Performance comparison in terms of RMSE between different regression methods to predict fingers and wrist movements from TMG data.}
	\label{Tab:TMGofflineMethodsRMSE}
	\footnotesize
	\renewcommand*{\arraystretch}{1.2}
	\begin{center}
		\setlength\tabcolsep{2.5pt}
		\begin{tabular}{|c|c|c||c|c|}
			\hline
			RR & ME & GP & TRR & TME\\
			\hline
			$0.45\pm 0.07$ & $0.33\pm 0.07$ & $0.30\pm 0.06$ & $0.35\pm0.11$ & $0.30\pm0.07$\\
			\hline
		\end{tabular}
	\end{center}
\end{table}
\normalsize

Table~\ref{Tab:TMGofflineMethodsCompTime} shows the average testing computation time for the tested regression methods. The computation times were measured using a non-optimized Python code on a laptop with 2.7GHz CPU and 32 GB of RAM.\footnote{Note that the structure of the proposed multilinear algebra problem allows numerical computation to be substantially improved in a number of ways.} 
We observe a testing time of $1$ ms for TME, which is reasonable for real-time applications allowing predictions at a frequency $>50$ Hz, as usually targeted by real-time detection of hand movements. Importantly, as opposed to GP, the computation testing time of TME is independent of the number of training data and depends only on the number of experts. Therefore, TME can be adapted to real-time predictions independently on the number of provided training data.
During training, TME converged with less than 10 iterations of the EM algorithm for all the participants, with a total training time of several minutes. This is mainly due to the fact that a TLR model is optimized at each step of the EM algorithm. While the training time of TME could be easily improved by using dedicated tensor libraries such as \cite{Kossaifi19}, and despite we used a naive implementation in our experiments, the training time remained reasonable for the method to be applied in real-time, as we show in the next subsection.

\begin{table}[t]
	\renewcommand*{\arraystretch}{1.2}
	\caption{Average testing computation time for the different regression methods. The methods are trained on $\sim 1000$ data samples. The testing computation time is measured for one data sample.}
	\label{Tab:TMGofflineMethodsCompTime}
	\begin{center}
		\begin{tabular}{c|c|c|c|c|c|}
			& RR & ME & GP & TRR & TME\\
			\hline
			Testing [ms] &$0.0065$ & $0.08$ &$1.5$& $0.035$& $1.0$\\
			\hline
		\end{tabular}
	\end{center}
\end{table}

\subsection{Real-time teleoperation with tactile myography}
\label{subsec:ExperimentTeleoperation}
To evaluate our method in a scenario closer to the end-user case, we conducted a real-time teleoperation experiment in which 11 non-amputated participants (one female and ten males) controlled a robotic hand and arm based on the activation of the muscles on their forearm. 

In the first part of the experiment, a protocol similar to the data collection of the experiment of Section~\ref{subsec:ExperimentTmg} was applied to collect TMG data associated with the hand postures of the participants. The tactile bracelet was placed on the forearm of the participant with the closing gap on the ulna bone. The participant, wearing the tactile bracelet and sitting in front of a monitor, was asked to replicate the movements of a model of the 24-DOFs dexterous motor hand of the Shadow Robot Company \cite{ShadowhandWebsite}. Similarly to the previous experiment, ground truth was obtained from the values of the animated hand model. Each participant executed four times the sequence of four movements, namely wrist flexion, wrist extension, power grasp and fingers extension. The participants were asked to perform the different movements in a relaxed way (particularly, the fingers were relaxed during wrist movements). Each stimulus follows a cycle of 14 s composed of a transition phase (2 s), an activation phase (6 s), a transition phase (2 s) and a relaxing phase (4 s). The data collected during the activation and relaxing phases, i.e., at zero and full activations, were used to train the regression models.

\begin{figure}[tbp]
	\centering
	\begin{subfigure}[b]{0.235\textwidth}
		\centering
		\includegraphics[width=\textwidth]{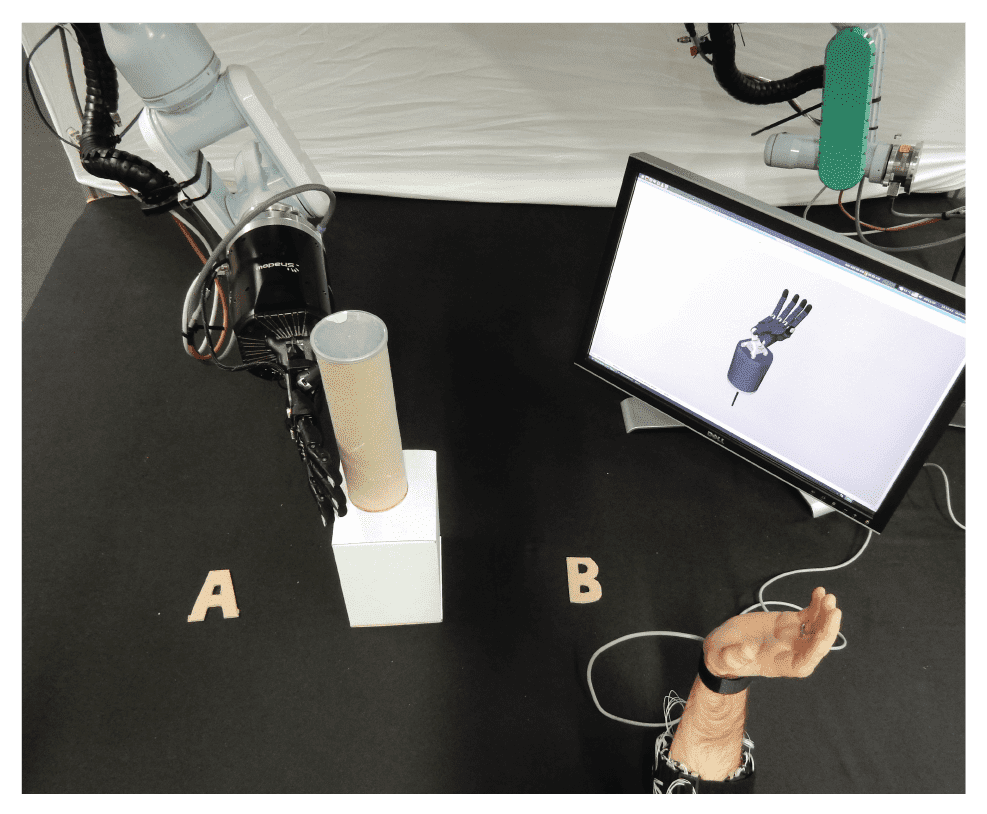}
		\caption{}
		\label{subFig:TeleopSetup}
	\end{subfigure}
	\begin{subfigure}[b]{0.235\textwidth}
		\centering
		\includegraphics[width=\textwidth]{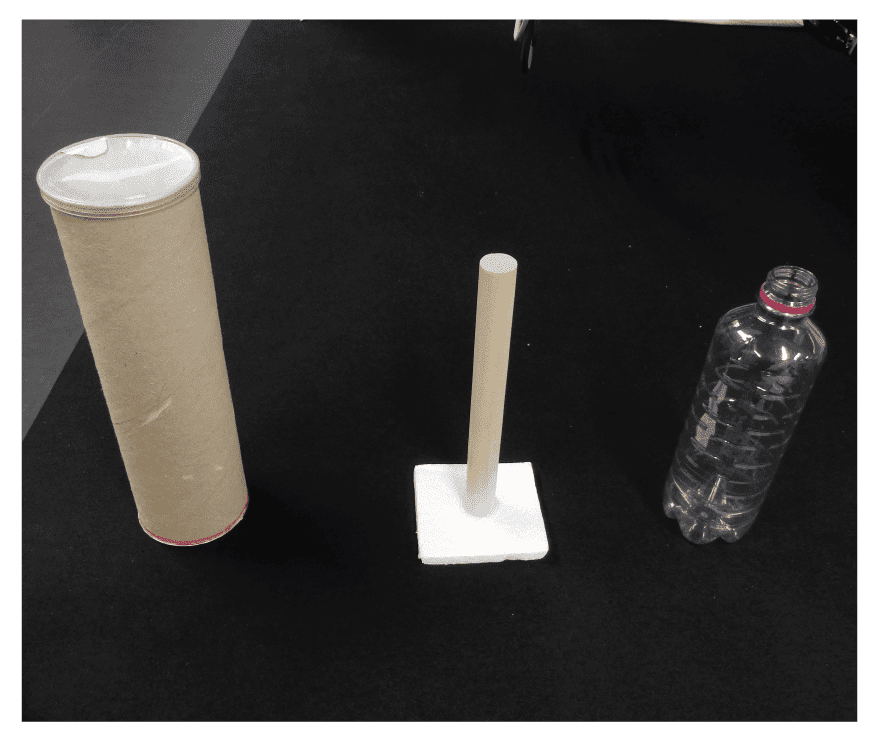}
		\caption{}
		\label{subFig:TeleopObjects}
	\end{subfigure}
	\caption{(\emph{a}) Setup of the teleoperation experiment. The participant is requested to grasp the chips cylinder and to place it on \emph{A} or \emph{B}. The distance between the initial position of the object and \emph{A} or \emph{B} is approximately $15$ cm. The detected activations are displayed on the monitor by an animated hand model. (\emph{b}) Objects used during the teleoperation experiment. }  
	\label{Fig:TeleopSetup}
\end{figure}

During the second part of the experiment, the participant teleoperated a Shadow robot hand mounted on a 7-DOFs Mitsubishi PA10 robot arm. 
S/he was sitting in front of the robotic system with the palm of the Shadow robot hand facing right, as showed in Figure~\ref{subFig:TeleopSetup}.
The different movements taught to the model in the first part of the experiments were mapped to the robotic system as follows: wrist flexion and extension were used to move the arm forward (in the direction of the palm) and backward (in the direction of the back of the hand), respectively. Power grasp and fingers extension were used to close and open the Shadow robot hand. When wrist flexion or extension was detected above a certain activation threshold, the velocity of the robot arm was incremented in the corresponding direction proportionally to the detected activation. Similarly, the posture of the robotic hand was incremented proportionally to the activation of power grasp or fingers extension if they were detected above a predefined threshold. The detected activations were also displayed on the Shadow robot hand model as in the first part of the experiment.

At the beginning of the second part of the experiment, the participant could get used to the learned mapping by controlling the simulated Shadow robot hand for a few minutes. Then, while teleoperating the real robotic system, the participant was asked to control the arm in order to approach it close to an object placed on a cube, to grasp this object and to bring it to a specific location on the left (\emph{A}) or on the right (\emph{B}) side and to release it. The complete setup is showed in Figure~\ref{subFig:TeleopSetup}. Three objects with different diameters were considered, namely a chips cylinder ($\diameter 75$ mm), a thin woodstick ($\diameter 21$ mm) and a PET bottle ($\diameter 63$ mm), as shown in Figure~\ref{subFig:TeleopObjects}. A total of 8 tasks were executed by each participant. The first six tasks consisted of bringing each object to \emph{A} and then to \emph{B}. Once a contact with the object was detected by the tactile fingertip sensors of the Shadow robot hand, the grasp pose was automatically maintained by the hand so that the subject could relax his/her fingers and focus on the wrist motion to steer the arm. The grasp pose was released as soon as a fingers extension command was detected. The maintenance of the grasp and the release were announced verbally by the system. The two last tasks consisted of bringing the PET bottle to \emph{A} and to \emph{B} without any holding assistance by the robot system. The time to complete each task was limited to two minutes. In case the object felt from the cube or was released out of the desired area, the experimenter replaced it at the initial position and the participant continued to execute the task in the remaining time. In case the participant could not control both the arm and the hand, e.g., if the arm was drifting continuously in one direction, the arm commands were disabled and the participant was requested to maintain a grasp on the object for 10 s before releasing it. Each participant tried to complete the 8 tasks with two different regression methods, namely TME and RR, trained on the data collected in the first part of the experiment. RR was chosen for comparison since it is considered as the baseline method for regression with TMG data. The order in which the two methods were tested was alternated between the participants.

\begin{figure}[tbp]
	\centering
	\includegraphics[width=0.235\textwidth]{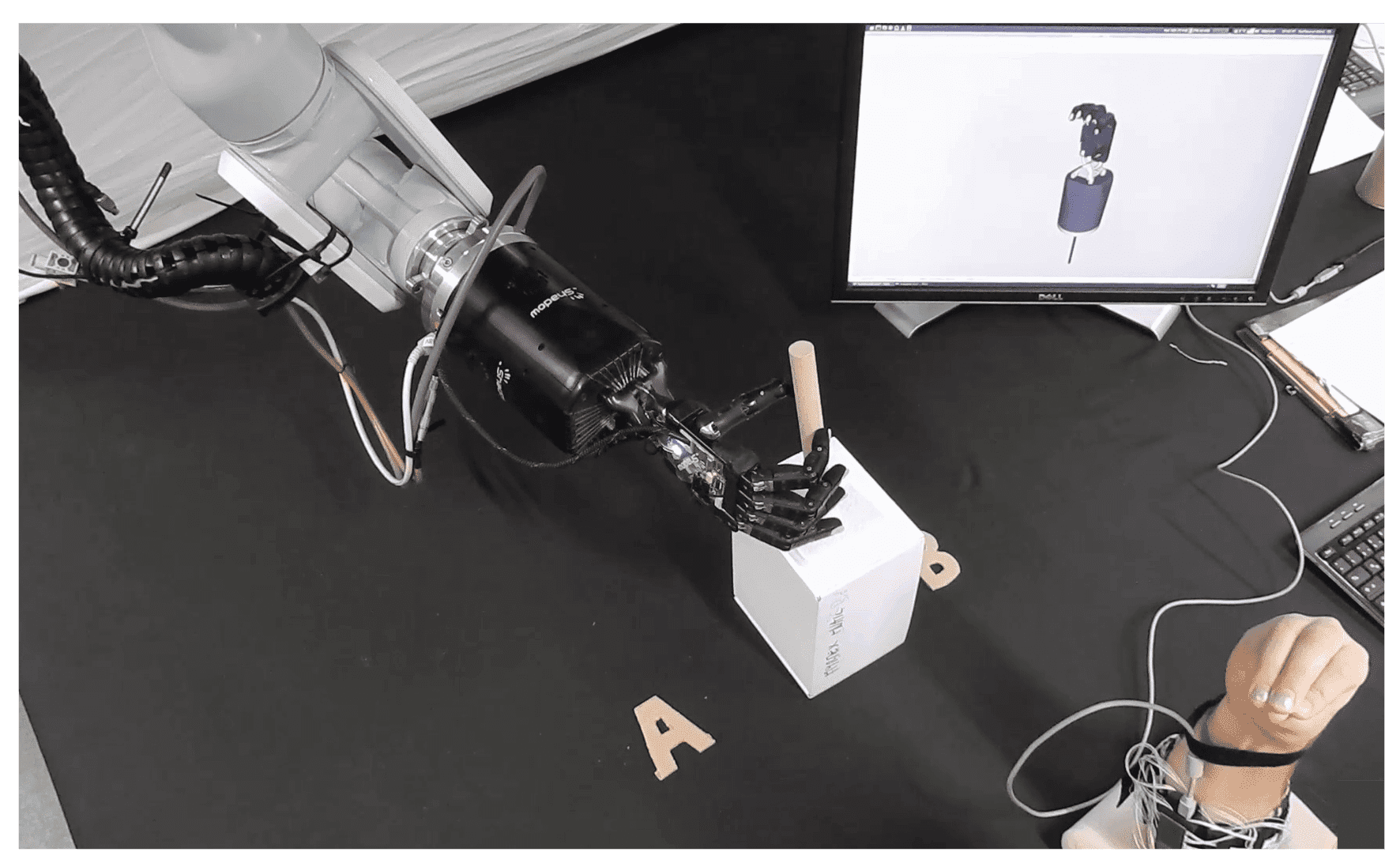}
	\includegraphics[width=0.235\textwidth]{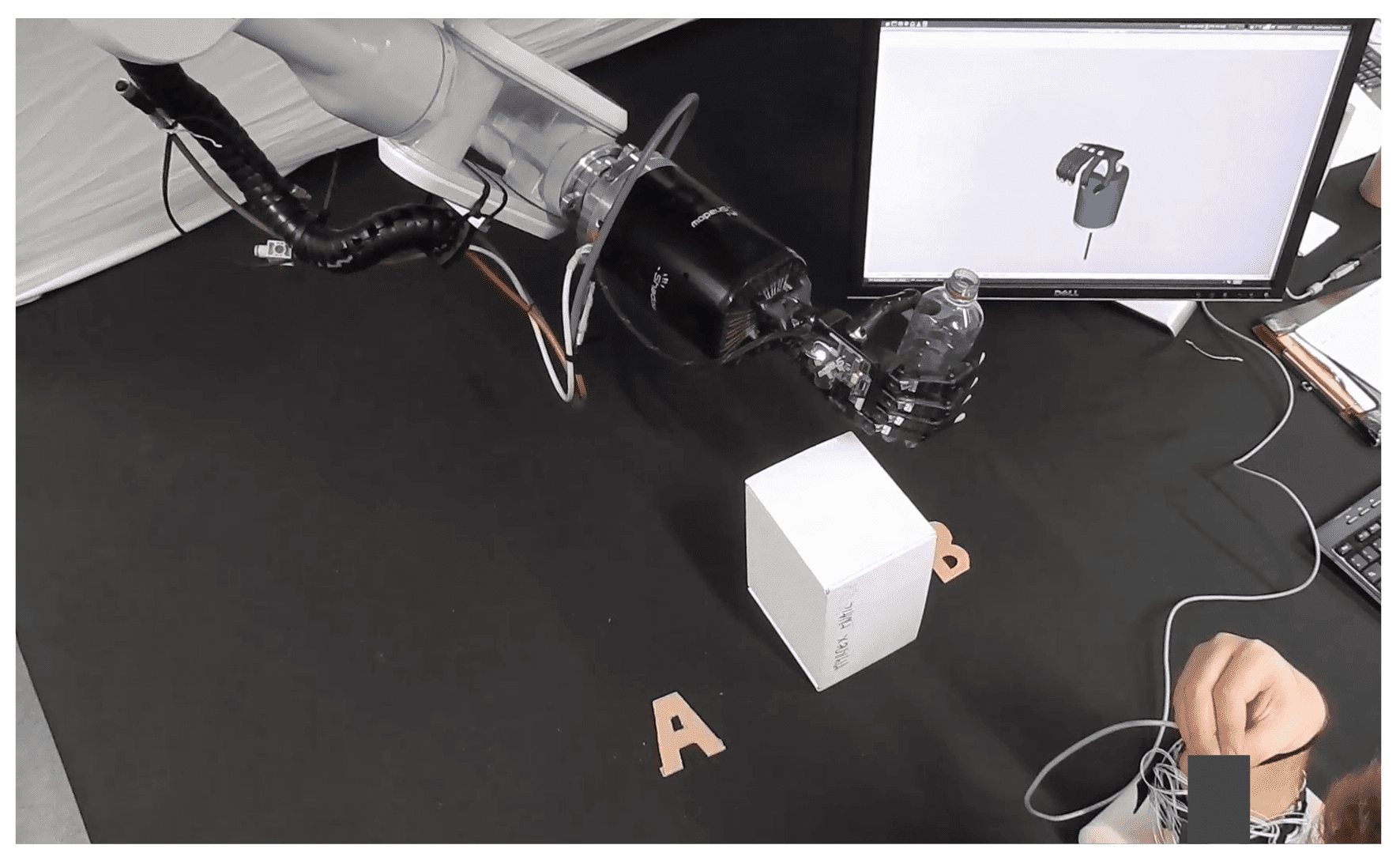}
	\includegraphics[width=0.235\textwidth]{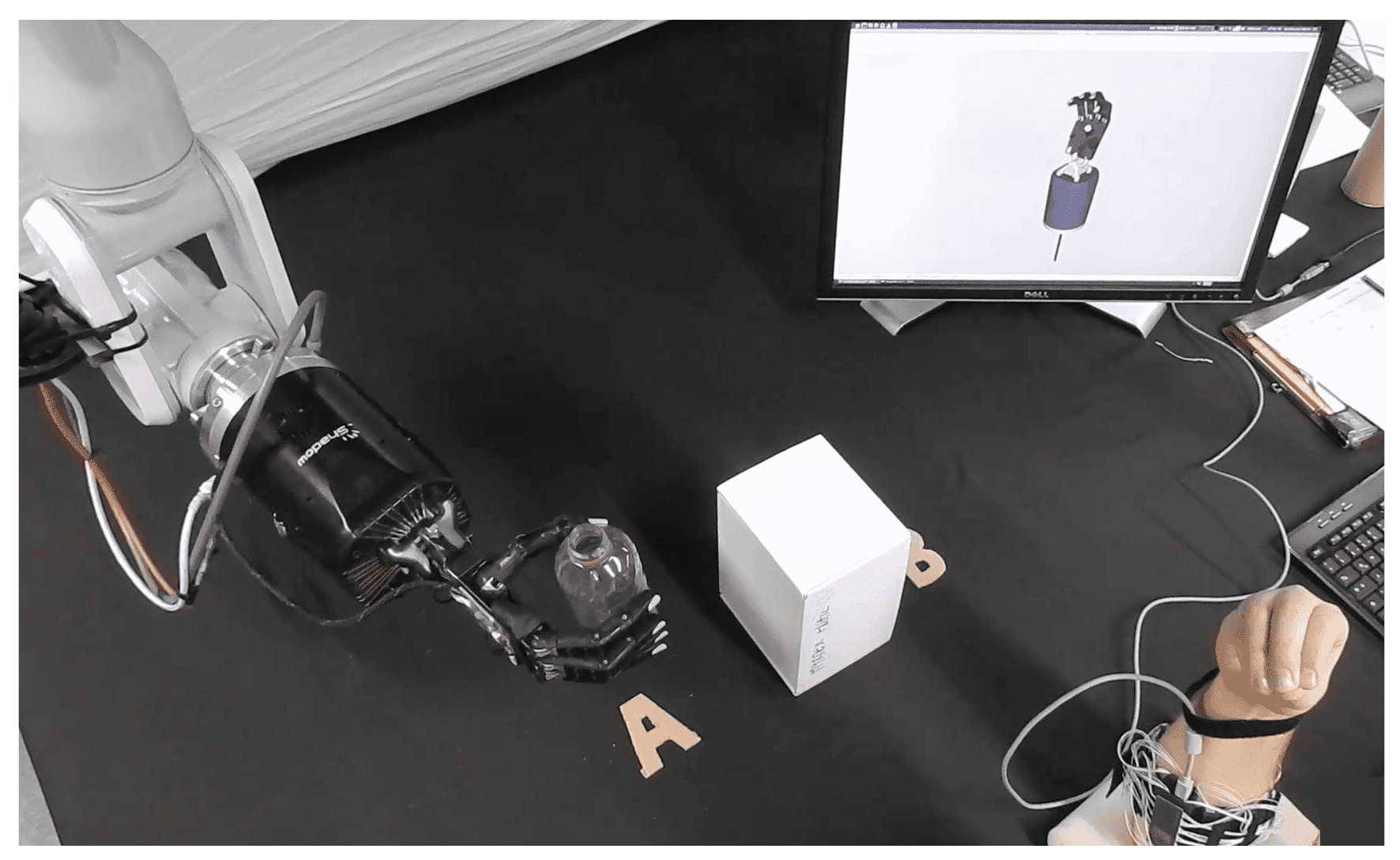}
	\includegraphics[width=0.235\textwidth]{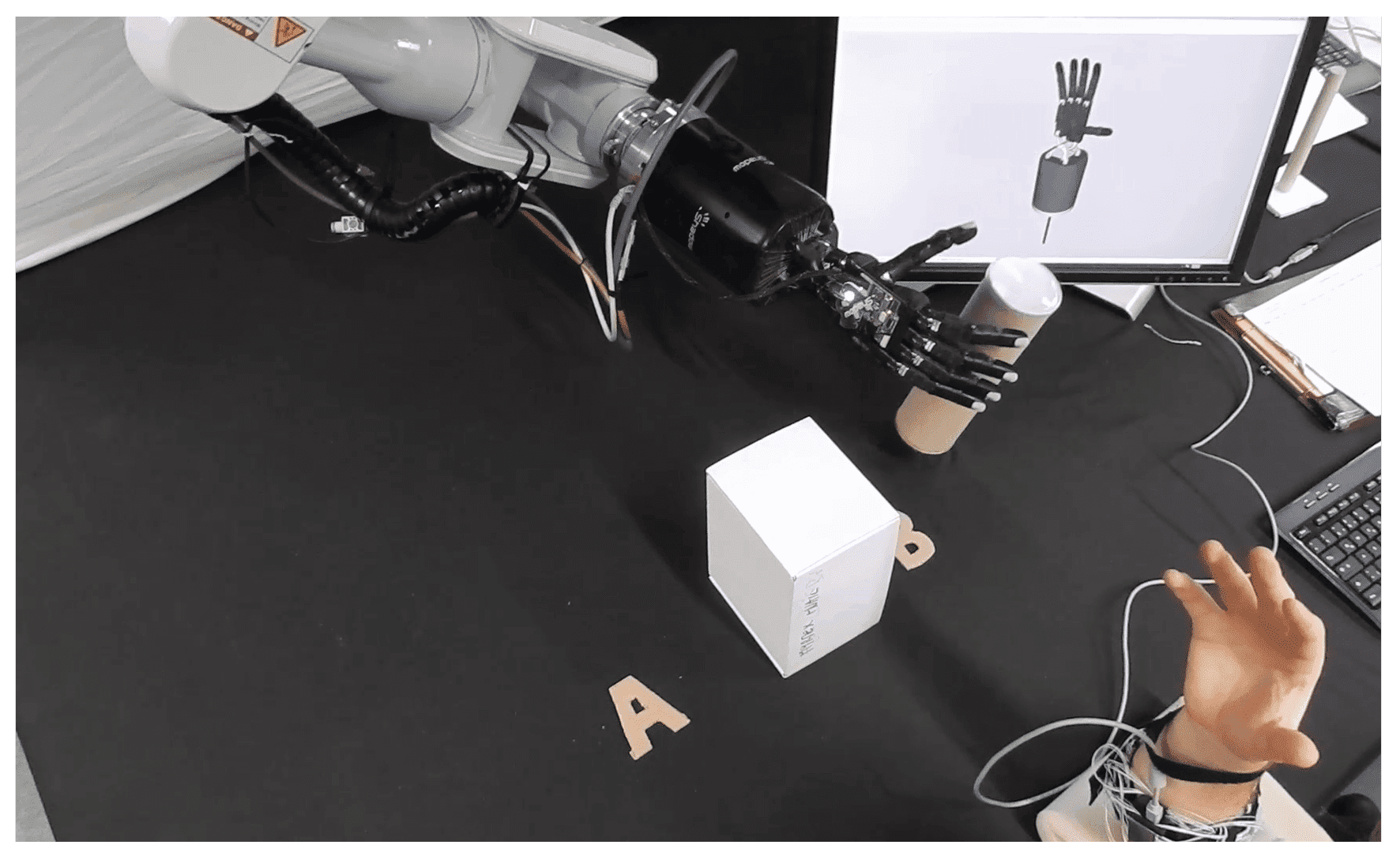}
	\caption{Snapshots of the teleoperation experiment for different tasks. A participant is (\emph{a}) grasping the woodstick, (\emph{b}) bringing the bottle to \emph{B}, (\emph{c}) reaching \emph{A} while holding the bottle, (\emph{d}) releasing the chips cylinder at \emph{B}. The detected activations are displayed by a Shadow robot hand model on the monitor.}  
	\label{Fig:TeleopResults}
\end{figure}

Figure~\ref{Fig:TeleopResults} shows snapshots of a participant executing different tasks. 6, respectively 7, out of the 11 participants were able to control both the arm and the hand during the whole experiment by using TME and RR, respectively. Note that the cases during which the arm commands had to be disabled occurred mainly for the second tested method (4 participants out of 5 testing TME as second method and 3 out of 4 participants testing RR as second method), suggesting a decrease of performance over time. One participant was not able to control both arm and hands for both methods and an other participant was able to control them for the 4 first tasks of the first tested method only.

The success rates, or ratios of successful tasks, for the case in which the participants controlled both robot arm and hand are presented in Table~\ref{Tab:TeleopSuccessObjects}. We observe that TME outperformed the performance RR by $~15\%$ when all the objects and both locations \emph{A} and \emph{B} are considered. The time needed to accomplish successful tasks were $55.6\pm 31.1$ s and $53.9\pm31.5$ s for TME and RR, respectively, showing almost no difference between the two methods.

For both methods, the tasks involving the woodstick result in the lowest success rate. Due to the small diameter of this object, the arm had to be positioned very precisely and a complete grasp activation had to be detected in order to perform a successful grasp. Therefore, the tasks involving the woodstick were the hardest to complete for the participants. In the case of TME, the success rate for the chips cylinder is lower than for the PET bottle. This can be explained by the fact that a small activation of the grasp movement was sometimes detected when the participants were flexing their wrist to make the arm move in the direction of the object. However, the robotic hand had to be completely opened to be able to be placed around the chips cylinder before grasping, while it could still be placed around the bottle if a small grasping activation was detected. In the case of RR, the success rate diminishes for the bottle compared to the chips cylinder. This may suggest a stronger decrease of performance over time with this method.

We observe that the success rates for the bottle are similar with and without the holding assistance activated for both methods. This result is particularly interesting as it shows that combinations of hand and wrist movements in this experiment, namely grasping with wrist flexion, and grasping with wrist extension, can be detected while training only on individual movements. Moreover, both aforementioned combinations were equally detected as the number of completed tasks for each location \emph{A} or \emph{B} was similar, i.e., 4 and 3 successful tasks for \emph{A} and \emph{B} with TME and 2 for each location with RR. Moreover, some of the participant did not wait that the contact with the object was detected before bringing it to its final location. Therefore, they managed to complete other tasks without using the holding assistance.

\begin{table}[t]
	\renewcommand*{\arraystretch}{1.2}
	\caption{Success rates over all the tasks and for each object in the case where the participants teleoperated both the robot arm and hand. The success rates are given in percent [$\%$].}
	\label{Tab:TeleopSuccessObjects}
	\footnotesize
	\begin{center}
		\setlength\tabcolsep{3.5pt}
		\begin{tabular}{c|c|c|c|c|c|}
			& Total & Chips cylinder & Woodstick & Bottle & Bottle (no HA)\\
			\hline
			TME & $45.8$ &$41.7$ &$16.7$ &$66.7$ &$58.3$ \\
			\hline
			RR & $30.6$ & $44.4$ &$12.5$& $35.7$& $28.6$\\
			\hline
		\end{tabular}
	\end{center}
\end{table}
\normalsize

Table~\ref{Tab:TeleopFailures} shows the proportion of the failed tasks for which the time ran out during each of the task steps, namely grasping, moving and releasing the object, for TME and RR. We observe that the proportions are similar for both methods with the grasping step being the main cause of failure, followed by the moving step. Failures during grasping occurred mainly because the detected grasp activation was not sufficient to grasp the object or when it was activated too soon, therefore resulting in the object being pushed out of the support box. Failures while moving were due to difficulties to detect wrist flexion and extension or to the object falling down while the participant was trying to reach \emph{A} or \emph{B}. Finally, if the fingers extension movement was not detected properly while the holding assistance was activated, the opening of the hand was not triggered, resulting in failure to release the object. 

\begin{table}[t]
	\renewcommand*{\arraystretch}{1.2}
	\caption{Proportion of the failed tasks due to running out of time during the grasp, the displacement and the release of the object [$\%$].}
	\label{Tab:TeleopFailures}
	\footnotesize
	\begin{center}
		\begin{tabular}{c|c|c|c|}
			& Grasping & Moving & Releasing\\
			\hline
			TME & $57.7$ &$30.8$ &$11.5$ \\
			\hline
			RR & $58.1$ &$32.6$& $9.3$\\
			\hline
		\end{tabular}
	\end{center}
\end{table}
\normalsize

The success rates for the case in which the arm commands where disabled and the participant was only requested to grasp the object are presented in Table~\ref{Tab:TeleopSuccessGraspOnly}. In this case, RR outperforms TME, especially for the tasks involving the woodstick. This can be explained by the fact that, in most of the cases, the arm commands had to be disabled for RR because the arm was drifting on the left or on the right, while it had to be disabled for TME because no activation was detected to move the arm, so that the participant could not position it to grasp the object. Generally, the detected activation of the grasp movement was also limited, therefore some tasks were difficult to achieve, especially those involving the woodstick.

\begin{table}[t]
	\renewcommand*{\arraystretch}{1.2}
	\caption{Success rates over all the tasks and for each object, for the case in which the commands of the arm had to be disabled. The success rates are given in percent [$\%$].}
	\label{Tab:TeleopSuccessGraspOnly}
	\setlength\tabcolsep{3.5pt}
	\footnotesize
	\begin{center}
		\begin{tabular}{c|c|c|c|c|c|}
			& Total & Chips cylinder & Woodstick & Bottle & Bottle (no HA)\\
			\hline
			TME & $38.8$ &$40.0$ &$0$ &$40.0$ &$40.0$ \\
			\hline
			RR & $53.8$ & $50.0$ &$66.7$& $50.0$& $50.0$\\
			\hline
		\end{tabular}
	\end{center}
\end{table}
\normalsize

\section{Discussion}
\label{sec:Discussion}
The proposed TME model allows the structure of tensor-valued data to be taken into account in the regression problem. Overfitting can then be reduced, which is particularly important when only few tensor-valued training data are available.
We showed the effectiveness of the approach to detect hand movements from TMG data, outperforming the other tested methods in an offline experiment and allowing participants to teleoperate a robotic arm and hand in real-time. 

It is important to emphasize that the tensor methods (TRR and TME) systematically outperformed their vector counterparts (RR and ME) in our experiments. This confirms that tensor methods not only result in tractable algorithms, but also improve the performance of the learning process by accounting for the underlying of structure of the data. These results highlight the importance of exploiting tensorial representations for learning algorithms when the data are naturally represented by matrices or tensors.
Moreover, for both vector- and tensor-based algorithms, the mixture of experts offers the advantage of combining the predictions from experts whose models are specialized for different regions of the input space. For a wide category of problems involving non-stationary or piecewise continuous regression processes as those considered in our experiments, mixture of experts (i.e., ME and TME) naturally outperform simpler regression methods (i.e., RR and TRR). In this context, the proposed TME model efficiently combines the advantages of tensor methods and of mixture of experts, and may therefore be beneficial in various applications.
Notice that, although we did not consider TRR and ME in our real-time experiment for the sake of keeping a reasonable experience time for the participants, we expect the aforementioned observations with respect to RR and TME to also apply in this case.

Moreover, it is worth highlighting that, during the real-time experiment, the TME model was able to successfully detect intermediate and combination of activation, while trained only with zero and complete individual movements. Notably, participants managed to activate wrist flexion or extension along with power grasp. This indicates that a holding assistance may not be required. However, some participants reported that the holding assistance was helpful as a feedback indicating that the grasp was effective or to make them feel more comfortable while teleoperating the arm, as they could focus on one movement only.

A decrease of performance over time, indicated by the necessity of deactivating the arm commands while testing the second method, seemed to have occurred during the real-time experiment. Moreover, some participants reported that they felt that the control of the robotic arm and hand was harder to perform over time. Moreover, we qualitatively observed that this problem seemed to occur particularly for participants who were trying to apply high forces to execute the different movements. We hypothesize that this is due to small displacements of the TMG bracelet over time, inducing a shift of the testing data compared to the training data. This problem could be overcome by improving the placement of the bracelet and by adapting the model over time. Techniques such as covariate shift adaptation in the case of sEMG \cite{Vidovic16} could be investigated.

It is important to emphasize the fact that the participants were able to adapt in some extent to the predictions of the method. They slightly modified their hand movements in order to obtain the desired action of the robotic arm and hand. Therefore, we observe a form of active learning, where the method learned from the training data, while the participants learned from the method in order to achieve the desired performance.

In both experiments, the ranks of the experts were given by a common value. The performance of TME may be further improved by selecting a specific rank for each expert. However, to avoid increasing computation time, automatic rank selection procedures have to be investigated. In particular, the automatic rank selection presented in \cite{Guo12} could be exploited to determine the rank of the expert models. The suggested method uses a $\ell_{1,2}$ norm regularization and optimizes the model with iteratively reweighted least squares (IRLS) algorithm. This approach seems promising as the authors reported in their experiments that the automatic procedure provided the same rank as the one selected by cross-validation. 

Overall, we believe that many robotic applications could benefit from the use of tensor methods. While this paper illustrated their use for recognizing hand movements from tactile myography with the proposed TME model, tensor methods may also be exploited for other types of tensor-valued data, e.g., for applications in robotic perception where images can be viewed as matrices, and video streams as third-order tensors. Moreover, tensor methods may be exploited in robot skills learning to efficiently represent and exploit skill parameters. In this direction, Zhao \etal ~\cite{Zhao17} proposed to generalize robotic skills by exploiting tensor decomposition to extract task-agnostic information from skill parameters represented as third-order tensors. Following similar ideas, tensor methods may also be used to represent task parameters in multiple coordinate systems, or to determine latent representations of high-dimensional skill or control parameters represented in tensor form.

\section{Conclusion}
\label{sec:Conclusion}
This paper presented an extension of mixture of experts to tensor-valued data. Our method brings together the advantages and robustness of mixture models and tensor methods. Therefore, it allows an efficiently combination of predictions from experts specialized in different regions of the input space, while taking into account the structure of tensor-valued data in the soft space division and in the predictions of the experts. The data are efficiently exploited, so that a model trained with a small amount of training data is able to achieve good performances, while overcoming the overfitting problem. This is particularly important in robotics as the amount of training data is often small compared to the dimensionality of the data. The effectiveness of our model was illustrated with artificially generated data and in two experiments aiming at detecting hand movements by measuring the pressure induced by the muscles activity of the forearm with tactile myography. We showed that the testing computational time of the proposed model is low, due to a computational cost independent of the number of training data, therefore making it compatible with real-time robotic applications. 

Future work will investigate automatic rank selection procedures with the objective of automatically determining all the ranks of the model, therefore avoiding the use of cross-validation in the training process.
Moreover, we will investigate extensions of the proposed tensor-variate mixture-of-experts (TME) model to other applications in robotics, as well as to more complex models, such as hierarchical TME \cite{Yuksel12}. It is worth noting that the proposed TME model permits to incorporate structural information of the data as a special case of neural network. Extensions of this model could then also lead to interesting perspectives in the development of neural network structures for tensor data that would have better interpretability, that could be trained with smaller amount of data, and that would provide better generalization results by avoiding overfitting. 

\section*{Acknowledgement}
We thank Guillaume Walck for his significant help on the preparation and execution of the experiment on the real robotic setup and for his suggestions during the preparation of the manuscript. This work was supported by the Swiss National Science Foundation and the German Research Foundation (SNSF/DFG project TACT-HAND).


\bibliography{References}

\end{document}